%% file: main.tex
\documentclass[10pt]{article} 
\usepackage[accepted]{tmlr}

\input{math_commands.tex}

\usepackage{hyperref}
\usepackage{url}

\usepackage{amsmath, amssymb, amsfonts}
\usepackage{authblk}
\usepackage{hyperref}
\usepackage{index}
\usepackage{graphicx}

\usepackage[normalem]{ulem}

\makeatletter
\renewcommand*{\@fnsymbol}[1]{\ensuremath{\ifcase#1\or *\or \dagger\or \ddagger\or
   \mathsection\or \mathparagraph\or \|\or **\or \dagger\dagger
   \or \ddagger\ddagger \else\@ctrerr\fi}}
\makeatother

\title{Align and Distill: Unifying and Improving\\ 
 Domain Adaptive Object Detection}

\author{\name Justin~Kay$^1$, Timm~Haucke$^1$, Suzanne~Stathatos$^2$, Siqi~Deng$^{3,*}$, Erik~Young$^4$, Pietro~Perona$^{2,3}$, Sara~Beery$^{1,=}$, Grant~Van~Horn$^{5,=}$ \\ 
\addr $^{1}$MIT, \addr $^{2}$Caltech, $^{3}$AWS, \addr $^{4}$Skagit Fisheries Enhancement Group, \addr $^{5}$UMass Amherst  \email kayj@mit.edu \\ \addr $^{=}$Equal advising contribution, $^{*}$Work done outside AWS}

\def\month{03}  
\def\year{2025} 
\def\openreview{\url{https://openreview.net/forum?id=ssXSrZ94sR}} 

\usepackage{multirow}
\usepackage{xspace}
\usepackage{enumitem}
\usepackage{float}
\usepackage[para,symbol]{footmisc}
\usepackage[capitalise]{cleveref}

\newcommand*{\methodname}{ALDI\@\xspace}
\newcommand*{\methodbest}{ALDI++\@\xspace}

\usepackage{xcolor} 
\usepackage{colortbl}
\usepackage{soul}

\definecolor{paleviolet}{HTML}{CC99FF}
\definecolor{paleblue}{HTML}{A9D1F7}
\definecolor{menthol}{HTML}{B4F0A7}
\definecolor{paleyellow}{HTML}{FFFFBF}
\definecolor{paleorange}{HTML}{FFDFBE}
\definecolor{melon}{HTML}{FFB1B0}
\definecolor{Burnin}{HTML}{EDEDED}
\colorlet{T_src}{paleviolet}
\colorlet{T_tgt}{paleblue}
\colorlet{MtM}{menthol}
\colorlet{Postprocess}{paleyellow}
\colorlet{L_distill}{paleorange}
\colorlet{L_align}{melon}
\usepackage[export]{adjustbox}
\makeatletter
\renewcommand*{\@fnsymbol}[1]{\ensuremath{\ifcase#1\or *\or \dagger\or \ddagger\or
   \mathsection\or \mathparagraph\or \|\or **\or \dagger\dagger
   \or \ddagger\ddagger \else\@ctrerr\fi}}
\makeatother

\usepackage{cleveref}
\usepackage{booktabs}
\usepackage{caption}
\usepackage{subfig}

\RequirePackage{xspace}
\makeatletter
\DeclareRobustCommand\onedot{\futurelet\@let@token\@onedot}
\def\@onedot{\ifx\@let@token.\else.\null\fi\xspace}

\def\eg{\emph{e.g}\onedot} 

\def\ie{\emph{i.e}\onedot}

\makeatother

\begin{document}

\maketitle
\vspace{-10pt}

\begin{abstract}
\setcounter{footnote}{1}  
\vspace{-5pt}
Object detectors often perform poorly on data that differs from their training set. 
Domain adaptive object detection (DAOD) methods have recently demonstrated strong results on addressing this challenge.
Unfortunately, we identify systemic benchmarking pitfalls that call past results into question and hamper further progress:  
(a)~Overestimation of performance due to underpowered baselines, (b)~Inconsistent implementation practices preventing transparent comparisons of methods, and (c)~Lack of generality due to outdated backbones and lack of diversity in benchmarks. We address these problems by introducing: (1)~A unified benchmarking and implementation framework, Align and Distill (ALDI), enabling comparison of DAOD methods and supporting future development, (2)~A fair and modern training and evaluation protocol for DAOD that addresses benchmarking pitfalls, (3)~A new DAOD benchmark dataset, CFC-DAOD, increasing the diversity of available DAOD benchmarks, and (4)~A new method, \methodbest, that achieves state-of-the-art results by a large margin. \methodbest outperforms the previous state-of-the-art by +3.5 AP50 on Cityscapes $\rightarrow$ Foggy Cityscapes, +5.7 AP50 on Sim10k $\rightarrow$ Cityscapes (where ours is the \textit{only} method to outperform a fair baseline), and +0.6 AP50 on CFC-DAOD. 
ALDI and \methodbest are architecture-agnostic, setting a new state-of-the-art for YOLO and DETR-based DAOD as well without additional hyperparameter tuning.
Our framework\footnote{\href{https://github.com/justinkay/aldi}{github.com/justinkay/aldi}}, dataset\footnote{\href{https://github.com/visipedia/caltech-fish-counting/tree/main/CFC-DAOD}{github.com/visipedia/caltech-fish-counting}}, 
and method offer a critical reset for DAOD and provide a strong foundation for future research. 
\end{abstract}
\vspace{-12pt}

\section{Introduction}
\label{sec:intro}
\vspace{-5pt}

\textbf{The challenge of DAOD.} 
Modern object detector performance, though excellent across many benchmarks~\citep{lin2014microsoft,weinstein2021benchmark,weinstein2021general,bondi2018spot,schneider2018deep,rodriguez2011density}, 
often severely degrades when test data exhibits a distribution shift with respect to training data~\citep{oza2023unsupervised}. For instance, detectors do not generalize well when deployed in new environments in environmental monitoring applications~\citep{kay2022caltech,weinstein2021benchmark}. Similarly, models in medical applications perform poorly when deployed in different hospitals or on different hardware than they were trained~\citep{xue2023cross,guan2021domain}. Unfortunately, in real-world applications it is often difficult, expensive, or time-consuming to collect the additional annotations needed to address such distribution shifts in a supervised manner.
An appealing option in these scenarios is \textit{unsupervised domain adaptive object detection (DAOD)}, which attempts to improve detection performance when moving from a ``source'' domain (used for training) to a ``target'' domain (used for testing)~\citep{koh2021wilds,kalluri2023gnet} without the use of target-domain supervision. 

\noindent
\textbf{The current paradigm.}
The research community has established a set of standard benchmark datasets and methodologies that capture the deployment challenges motivating DAOD. Benchmarks consist of labeled data that is divided into two sets: a source and a target, each originating from different domains. DAOD methods are trained with source-domain images and labels, as in traditional supervised learning, and have access to unlabeled target domain images. The target-domain labels are not available for training. 

To measure DAOD methods' performance, researchers use \textit{source-only models} and \textit{oracle models} as points of reference. Source-only models---sometimes also referred to as \textit{baselines}---are trained with source-domain data only, representing a lower bound for performance without domain adaptation. Oracle models are trained with \textit{supervised} target-domain data, representing a fully-supervised upper bound. The goal in DAOD is to close the gap between source-only and oracle performance without target-domain supervision.

\begin{figure}
    \centering
    \begin{minipage}[c]{.25\linewidth}
      \centering
      \includegraphics[width=\linewidth]{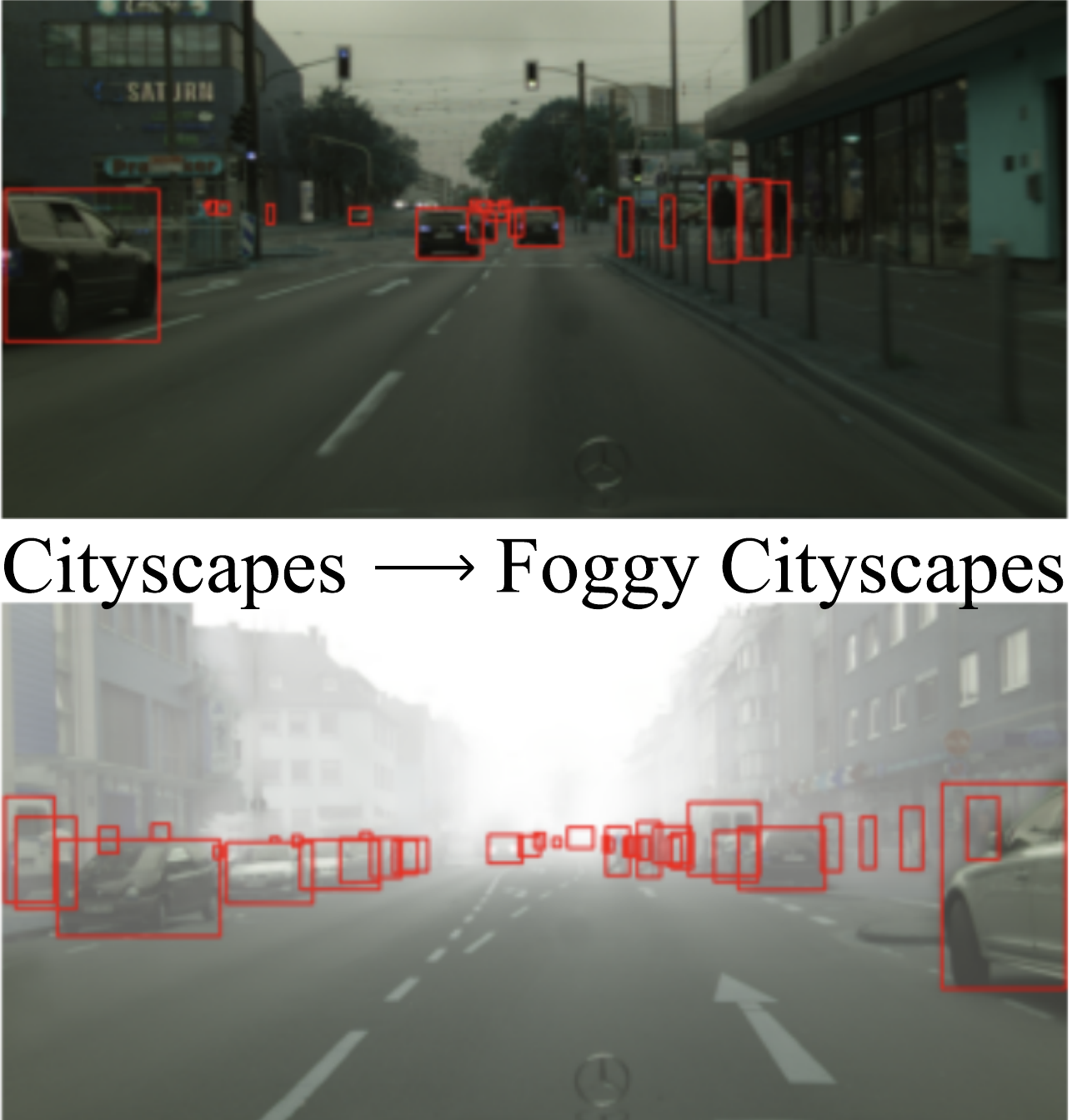}
    \end{minipage}%
    \begin{minipage}[c]{.75\linewidth}
      \centering
      \includegraphics[width=\linewidth]{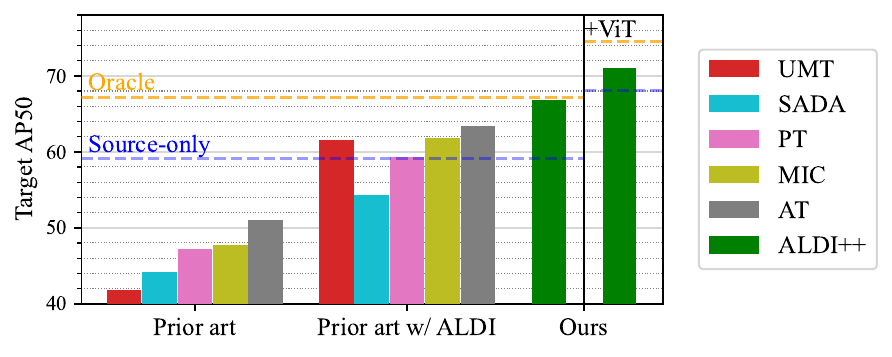} \\
    \end{minipage}
    \caption{\textbf{\methodname provides a unified framework for fair comparison of domain adaptive object detection (DAOD) methods, and ALDI++
    achieves state-of-the-art performance.}
    We show: 
    \textbf{(1)}~Inconsistent implementation practices give the appearance of steady progress in DAOD (left bars \citep{deng2021unbiased,chen2021scale,chen2022learning,hoyer2023mic,li2022cross}); reimplementation and fair comparison with ALDI shows less difference between methods than previously reported (middle bars);
    \textbf{(2)}~A fairly constructed source-only model (blue line) outperforms many existing DAOD methods, indicating less progress has been made than previously reported; and a proper oracle (orange line) outperforms \textit{all} existing methods, in contrast to previously-published results; 
    and \textbf{(3)}~Our proposed method \methodbest (green bars)
    achieves state-of-the-art performance on  DAOD benchmarks such as Cityscapes~$\rightarrow$~Foggy Cityscapes and is complementary to ongoing advances in object detection like VitDet~\citep{li2022exploring}.
    }
    \label{fig:fig1}
\end{figure}

\noindent
\textbf{Impediments to progress.}
Recently-published results indicate DAOD is exceptionally effective, doubling the performance of source-only models and even outperforming fully-supervised oracles~\citep{li2022cross,chen2022learning,cao2023contrastive}.
However, upon close examination we discover problems with current benchmarking practices that call these results into question:

\noindent 
\ul{P1: Improperly constructed source-only and oracle models, leading to overestimation of performance gains.} 
We find that source-only and oracle models are consistently constructed in a way that does not properly isolate domain-adaptation-specific components, 
leading to misattribution of performance improvements. 
We show that when source-only and oracle models are fairly constructed---\ie use the same architecture and training settings as DAOD methods---no existing methods outperform oracles and \textit{many methods do not even outperform source-only models} (\cref{fig:fig1}), in stark contrast to claims made by recent work. 
These results mean we do not have an accurate measure of the efficacy of DAOD.

\noindent
\ul{P2: Inconsistent implementation practices preventing transparent comparisons of methods.}
We find existing DAOD methods are built using a variety of different object detection libraries with inconsistent training settings, making it difficult to determine whether performance improvements come from new DAOD methods or simply improved hyperparameters. We find that tweaking these hyperparameters---whose values often differ between methods yet are not reported in papers---can lead to a \textit{larger change in performance than the proposed methods themselves} (see \cref{sec:ablations}), thus we cannot take reported advancements at face value.
Without the ability to make fair comparisons we cannot transparently evaluate contributions nor make principled methodological progress. 

\noindent
\ul{P3: (a) Lack of diverse benchmarks and (b) outdated backbone architectures, leading to overestimation of methods' generality.}
DAOD benchmarks have focused largely on urban driving scenarios with synthetic distribution shifts~\citep{sakaridis2018semantic,johnson2016driving}, and methods continue to use outdated detector backbones for comparison with prior work~\citep{chen2018domain}. We show that in fact the \textit{ranking of methods changes across benchmarks and architectures}, revealing that published results may be uninformative for practitioners using modern architectures and real-world data.

\noindent
\textbf{A critical reset for DAOD research.}
DAOD has the potential for impact in a range of real-world applications, but these systemic benchmarking pitfalls impede progress.
We aim to address these problems and lay a solid foundation for future progress in DAOD with the following contributions:

\noindent
\ul{1. \textit{Align and Distill (ALDI)}, a unified benchmarking and implementation framework for DAOD.} In order to enable fair comparisons, we first identify key themes in prior work (\cref{sec:related}) and unify common components into a single state-of-the-art framework, \textit{ALDI} (\cref{sec:framework}). 
ALDI facilitates detailed study of prior art and streamlined implementation of new methods,
supporting future research.

\noindent
\ul{2. A fair and modern training protocol for DAOD methods, enabled by \methodname.} We provide quantitative evidence of the benchmarking pitfalls we identify and propose an updated training and evaluation protocol to address them (\cref{sec:baselinesoracles}). This enables us to set more realistic and challenging targets for the DAOD community and perform the first fair comparison of prior work in DAOD (\cref{sec:fair_comparison}).

\noindent
\ul{3. A new benchmark dataset, CFC-DAOD}, sourced from a real-world adaptation challenge in environmental monitoring (\cref{sec:data}).
CFC-DAOD increases the diversity of DAOD benchmarks and is notably larger than existing options. We show that the ranking of methods changes across different benchmarks (\cref{sec:fair_comparison}), thus the community will benefit from an additional point of comparison.

\noindent
\ul{4. A new method, \methodbest, that achieves state-of-the-art results by a large margin.}
Using the same model settings across all benchmarks, \methodbest outperforms the previous state-of-the-art by +3.5 AP50 on Cityscapes $\rightarrow$ Foggy Cityscapes, +5.7 AP50 on Sim10k $\rightarrow$ Cityscapes (where ours is the \textit{only} method to outperform a fair source-only model), and +2.0 AP50 on CFC Kenai $\rightarrow$ Channel.

\section{Related Work}
\label{sec:related}

\noindent
Our work concerns domain adaptive 2D object detection (DAOD). Two methodological themes have dominated recent DAOD research: \textit{feature alignment} and \textit{self-training/self-distillation}. We first give an overview of these themes and previous efforts to combine them, and then use commonalities to motivate our unified framework, \textit{Align and Distill}, in \cref{sec:framework}. 

\noindent
\textbf{Feature alignment in DAOD.} Feature alignment methods aim to make target-domain data ``look like'' source-domain data, reducing the magnitude of the distribution shift. The most common approach utilizes an adversarial learning objective to align the feature spaces of source and target data~\citep{ganin2015unsupervised,chen2021scale,chen2018domain,zhu2019adapting}.
Faster R-CNN in the Wild~\citep{chen2018domain} utilizes adversarial networks at the image and instance level. SADA~\citep{chen2021scale} extends this to multiple adversarial networks at different feature levels. Other approaches propose mining for discriminative regions~\citep{zhu2019adapting}, weighting local and global features differently~\citep{saito2019strong}, incorporating uncertainty~\citep{nguyen2020domain}, and using attention networks~\citep{vs2021mega}.
Alignment at the pixel level has also been proposed using image-to-image translation techniques to modify input images directly~\citep{deng2021unbiased}. 

\noindent
\textbf{Self-training/self-distillation in DAOD.} Self-training methods use a ``teacher'' model to predict pseudo-labels on target-domain data that are then used as training targets for a ``student'' model. 
Self-training can be seen as a type of \textit{self-distillation}~\citep{pham2022revisiting,caron2021emerging}, which is a special case of knowledge distillation~\citep{hinton2015distilling, chen2017learning} where the teacher and student models share the same architecture. 
Most recent self-training approaches in DAOD are based on the Mean Teacher~\citep{tarvainen2017mean} framework, in which the teacher model is updated as an exponential moving average (EMA) of the student model's parameters. 
Extensions to Mean Teacher for DAOD include: MTOR, which utilizes graph structure to enforce student-teacher feature consistency~\citep{cai2019exploring}, Probabilistic Teacher (PT), which uses probabilistic localization prediction and soft distillation losses~\citep{chen2022learning}, and Contrastive Mean Teacher (CMT), which uses MoCo~\citep{he2020momentum} for student-teacher consistency~\citep{cao2023contrastive}. 

\noindent
\textbf{Combining feature alignment and self-training.} Several approaches utilize \textit{both} feature alignment and self-training/self-distillation, motivating our unified framework. Unbiased Mean Teacher (UMT)~\citep{deng2021unbiased} uses mean teacher in combination with image-to-image translation to align source and target data at the pixel level. Adaptive Teacher (AT)~\citep{xue2023cross} uses mean teacher with an image-level discriminator network. Masked Image Consistency (MIC)~\citep{hoyer2023mic} uses mean teacher, SADA, and a masking augmentation to enforce teacher-student consistency. 
Because these methods were implemented in different codebases using different training recipes and hyperparameter settings, it is unclear which contributions are most effective and to what extent feature alignment and self-training are complementary. We address these issues by reimplementing these approaches in the ALDI framework and perform fair comparisons and ablation studies in \cref{sec:results}.

\noindent
\textbf{DAOD implementations.} There are two components to an object detector design: the detection architecture (\textit{e.g.} Faster R-CNN~\cite{ren2015faster}, YOLO~\cite{redmon2016lookonceunifiedrealtime}, DETR~\cite{carion2020end}) and the backbone (\textit{e.g.} VGG~\cite{simonyan2014very}, ResNet~\cite{he2016deep}, ViT~\cite{dosovitskiy2020image}). Current state-of-the-art methods in DAOD predominantly use Faster R-CNN architectures. DOAD methods for YOLO and DETR backbones have recently received some attention~\cite{zhou2023ssda,yu2022mttrans,jia2023pm}, but have yet to surpass Faster R-CNN-based methods' performance. For this reason, our main experiments also utilize the Faster R-CNN architecture. Existing methods differ in their choice of backbone, making comparisons difficult; we address this by consistently utilizing ResNet-50 backbones for all experiments and in our re-implementations of prior work. However, the ALDI framework is architecture and backbone agnostic, and we provide additional experiments using YOLO and DETR architectures, as well as ViT and ConvNeXt~\cite{liu2022convnet} backbones.

\noindent
\textbf{DAOD datasets.}  Cityscapes (CS) $\rightarrow$ Foggy Cityscapes (FCS)~\citep{cordts2016cityscapes,sakaridis2018semantic} is a popular DAOD benchmark that emulates domain shift caused by changes in weather in urban driving scenarios. 
The dataset contains eight vehicle and person classes. Sim10k $\rightarrow$ CS~\citep{johnson2016driving} poses a Sim2Real challenge, adapting from video game imagery to real-world imagery. 
The benchmark focuses on a single class, ``car''. Other common tasks include adapting from real imagery in PascalVOC~\citep{everingham2010pascal} to clip art and watercolor imagery~\citep{inoue2018cross}. We report results on CS $\rightarrow$ FCS and Sim10k $\rightarrow$ CS due to their widespread popularity in the DAOD literature and focus on real applications. We note that existing benchmarks reflect a relatively narrow set of potential DAOD applications. To study whether methods generalize outside of urban driving scenarios, in \cref{sec:data} we introduce a novel dataset sourced from a real-world adaptation challenge in environmental monitoring, where imagery is much different from existing benchmarks.

\section{Align and Distill (ALDI): Unifying DAOD}
\label{sec:framework}

\begin{figure}
  \begin{minipage}[b]{0.40\linewidth}
    \centering
    \includegraphics[width=\linewidth]{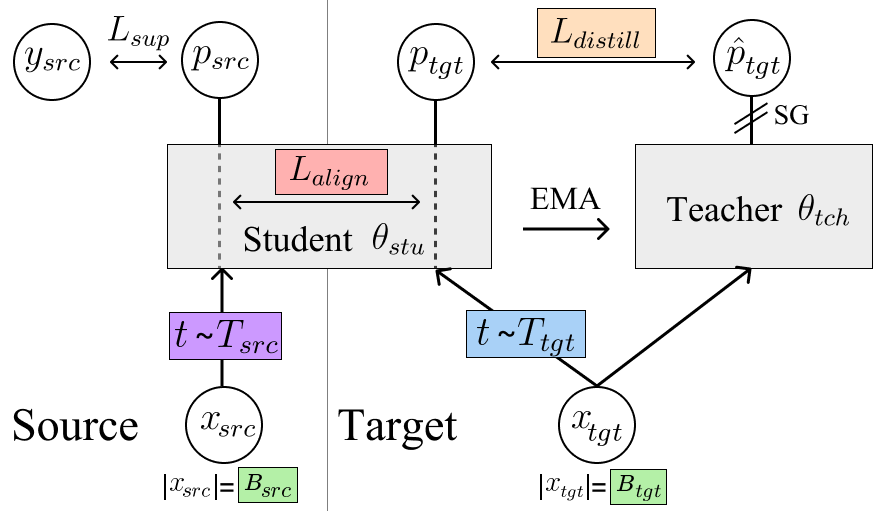}
    \par\vspace{0pt}
  \end{minipage}%
  \begin{minipage}[b]{0.60\linewidth}

\resizebox{\linewidth}{!}{
  
    \centering%
    \begin{tabular}{c|l}
    {\large ALDI module} & {\large Example settings} \\
    \hline\hline
    \rule{0pt}{\normalbaselineskip}\cellcolor{Burnin}Network init. & ImageNet \citep{chen2021scale,deng2021unbiased,hoyer2023mic}, \\
    \cellcolor{Burnin}{$\theta_{stu}, \theta_{tch}$} & Source-domain burn-in \citep{chen2022learning,li2022cross}\\
    
    \hline
    
    \rule{0pt}{\normalbaselineskip}\cellcolor{T_src}$T_{src}$, & Flip and scale, SimCLR augs \citep{chen2022learning,li2022cross},\\
    \cellcolor{T_tgt}$T_{tgt}$ & Image masking \citep{hoyer2023mic}\\

    \hline
    
    \rule{0pt}{\normalbaselineskip}\cellcolor{MtM} & $1:1$ \citep{chen2021scale,deng2021unbiased,hoyer2023mic}, \\
    \multirow{-2}{*}{\cellcolor{MtM}$B_{src}:B_{tgt}$} & $2:1$ \citep{chen2022learning,li2022cross} \\

    \hline



    \rule{0pt}{\normalbaselineskip}\cellcolor{L_distill} & 
    Pseudo-label
    \citep{deng2021unbiased,hoyer2023mic,li2022cross}, \\
    \multirow{-2}{*}{\cellcolor{L_distill}{$L_{distill}$}} & Knowledge distillation \citep{chen2022learning}\\

    \hline

    \rule{0pt}{\normalbaselineskip}\cellcolor{L_align} & Adversarial \citep{li2022cross,chen2021scale,hoyer2023mic}, \\
    \cellcolor{L_align} \multirow{-2}{*}{{$L_{align}$}} & Image-to-image translation \citep{deng2021unbiased}

    \end{tabular}%

}

\par\vspace{0pt}

\end{minipage}
\caption{\textbf{(Left) The \methodname 
framework.
} Each training step (moving left to right and bottom to top): \textbf{(1)} Sample $B_{\text{src}}$ labeled source images $x_{\text{src}}$; transform by $t \sim T_{\text{src}}$; pass to student; compute supervised loss $L_{\text{sup}}$ using ground-truth labels $y_{\text{src}}$. \textbf{(2)} Sample $B_{\text{tgt}}$ unlabeled target images $x_{\text{tgt}}$; transform by $t \sim T_{\text{tgt}}$; pass to student to get preds $p_{\text{tgt}}$. Compute alignment objectives $L_{\text{align}}$ using $x_{\text{src}}$ and $x_{\text{tgt}}$. \textbf{(3)} Pass same unlabeled target data $x_{\text{tgt}}$, weakly transformed, to teacher; postprocess to obtain teacher predictions $\hat{p}_{\text{tgt}}$. Compute distillation loss $L_{\text{distill}}$ between teacher and student predictions. Use stop gradient (SG) on teacher model; update teacher to the EMA of student's weights. \textbf{(Right)~Example settings for each component of \methodname.}  \methodname supports a range of existing methods off-the-shelf while providing a general implementation framework for new methods.}
\label{fig:framework}
\end{figure}

We first introduce \textit{Align and Distill (ALDI)}, a new benchmarking and implementation framework for DAOD.
ALDI unifies feature alignment and self-distillation approaches in a common framework, enabling fair comparisons and addressing \ul{P2. Inconsistent implementation practices}, while also
providing the foundation for development of a new method ALDI++ that achieves state-the-art performance (\cref{sec:ours}, \cref{sec:final_results}).
The framework is visualized in \cref{fig:framework}. All components are ablated in \cref{sec:ablations}.

\noindent
\textbf{Data.} DAOD involves two datasets: a labeled source dataset $X_{\text{src}}$ and an unlabeled target dataset $X_{\text{tgt}}$. Each training step, a minibatch of size $B$ is constructed containing both $B_{\text{\text{src}}}$ source images and $B_{\text{\text{tgt}}}$ target images, $B = B_{\text{src}} + B_{\text{tgt}}$.

\noindent
\textbf{Models.} ALDI is designed as a student-teacher framework to facilitate algorithms utilizing self-training/self-distillation. When enabled, both a student model \(\theta_{\text{stu}}\) and a teacher model \(\theta_{\text{tch}}\) are initialized with the same weights, typically obtained through supervised pretraining on ImageNet or \( X_{\text{src}} \). Pretraining on \( X_{\text{src}} \) is often referred to as ``burn-in.'' The student is trained via backpropagation, while the teacher's weights are updated each training step to be the EMA of the student's weights~\citep{tarvainen2017mean}, \ie $\theta_{\text{tch}} = \alpha \theta_{\text{tch}} + (1 - \alpha) \theta_{\text{stu}}$ with \(\alpha \in [0, 1]\). After training, we keep $\theta_{\text{tch}}$ and discard $\theta_{\text{stu}}$. Algorithms that do not use self-training/self-distillation (\eg SADA~\citep{chen2021scale}) simply disable $\theta_{\text{tch}}$.

In this paper we focus predominantly on two-stage detectors based on Faster R-CNN~\citep{ren2015faster} as they are currently the state-of-the-art in DAOD, though we note that our framework is architecture-agnostic and also supports YOLO and DETR-based detectors. We provide additional YOLO and DETR results in \cref{appendix:yolodetr}.

\noindent
\textbf{Training} involves one or more of the following three objectives. We note that each objective is optional in order to support a range of algorithmic approaches.

\noindent
\textbf{1. Supervised training with source data.} For each labeled source sample \( x_{\text{src},i} \), we apply a transformation \( t \sim T_{\text{src}} \), where \( T_{\text{src}} \) is the set of possible source-domain transformations. The transformed sample is passed through the student model to compute the supervised loss \( \mathcal{L}_{\text{sup}} \) given the ground truth targets \( y_{\text{src},i} \):

\begin{equation}
\mathcal{L}_{\text{sup}} = \frac{1}{B_{\text{src}}} \sum_{i=1}^{B_{\text{src}}} \mathcal{L}\left( \theta_{\text{stu}}(t(x_{\text{src},i})), y_{\text{src},i} \right)
\label{eq:sup}
\end{equation}

\noindent
where \(\mathcal{L}(\cdot, \cdot)\) are standard object detection loss functions, \eg those of Faster R-CNN~\citep{ren2015faster}.

\noindent
\textbf{2. Self-distillation with target data.} For each unlabeled target sample \( x_{\text{tgt},i} \), we transform the input using \(\hat{t} \sim T_{\text{weak}}\) (a set of weak transformations) for the teacher model and \( t \sim T_{\text{tgt}} \) (stronger transformations) for the student model. The teacher's predictions \(\hat{p}_{\text{tgt},i}\) serve as distillation targets for the student's predictions \( p_{\text{tgt},i} \), and we compute distillation loss \(\mathcal{L}_{\text{distill}}\) :

\vspace{-10pt}
\begin{minipage}{0.28\linewidth}
    \begin{equation}
        \hat{p}_{\text{tgt},i} = \theta_{\text{tch}}(\hat{t}(x_{\text{tgt},i}))
        \label{eq:2}
    \end{equation}
\end{minipage}
\begin{minipage}{0.28\linewidth}
    \begin{equation}
        p_{\text{tgt},i} = \theta_{\text{stu}}(t(x_{\text{tgt},i}))
        \label{eq:3}
    \end{equation}
\end{minipage}
\begin{minipage}{0.43\linewidth}
    \begin{equation}
    \mathcal{L}_{\text{distill}} = \frac{1}{B_{\text{tgt}}} \sum_{i=1}^{B_{\text{tgt}}} \mathcal{L}_{\text{distill}}\left( p_{\text{tgt},i}, \hat{p}_{\text{tgt},i} \right)
    \end{equation}
\end{minipage}

\noindent
where teacher outputs \(\hat{p}_{\text{tgt},i}\) are postprocessed to be either soft (\eg, logits or softmax outputs) or hard (\eg, thresholded pseudo-label) targets and the choice of distillation loss is method-specific.
This formulation unifies different distillation techniques into a common objective, supporting a range of approaches.

\noindent
\textbf{3. Feature alignment.} The source samples \( x_{\text{src},i} \) and target samples \( x_{\text{tgt},i} \) are optionally ``aligned'' using an alignment objective \( \mathcal{L}_{\text{align}} \) that enforces invariance across domains at either the image or feature level. This formulation is general; however, in this paper, we focus on two common alignment losses: domain-adversarial training and image-to-image alignment. 

Domain-adversarial training (\ie DANN~\citep{ganin2015unsupervised}) trains a domain classifier \( D \) to distinguish between source and target features, while the feature extractor aims to confuse \( D \):

\begin{equation}
\mathcal{L}_{\text{align,DANN}} = - \frac{1}{B} \sum_{i=1}^{B} \left[ y_{\text{dom},i} \log(D({\theta}(x_i))) + (1 - y_{\text{dom},i}) \log(1 - D({\theta}(x_i))) \right]
\end{equation}

where \( y_{\text{dom},i} \) is the domain label (source = 0, target = 1) and \({\theta}(x_i) \) is a feature representation of \( x_i \).

Image-to-image alignment instead pursues domain invariance in the pixel space. Given image-to-image generative models \( G_{\text{src}} \), \( G_{\text{tgt}} \) (\eg, a CycleGAN \citep{zhu2017unpaired}), images are ``translated'' (a pixel-level transformation) from the source domain to the target domain and vice versa. We then obtain $x_{\text{tgt-like},i} = G_{\text{src}}\left(x_{\text{src},i}\right)$, $x_{\text{src-like},i} = G_{\text{tgt}}\left(x_{\text{tgt},i}\right)$, and substitute into \cref{eq:sup}, \cref{eq:2}, and \cref{eq:3}.

\noindent
\textbf{Unification of prior work.} We demonstrate the generality of our framework by reimplementing five recently-proposed methods on top of ALDI for fair comparison: 
UMT~\citep{deng2021unbiased}, SADA~\citep{chen2021scale}, 
PT~\citep{chen2022learning}, MIC~\citep{hoyer2023mic}, and AT~\citep{li2022cross}. 
We enumerate the settings required to reproduce each method in \cref{appendix:experiment_settings}.

\section{\methodbest: Improving DAOD}
\label{sec:ours}

We next propose a set of simple but effective enhancements to the \textit{Align and Distill} approach. We call the resulting method \methodbest. We show in \cref{sec:final_results} that these enhancements lead to state-of-the-art results, and ablate each component in \cref{sec:ablations}.

\noindent
\textbf{1. Robust burn-in.} 
A key challenge in student-teacher methods is improving target-domain pseudo-label quality. 
We point out that pseudo-label quality in the early stages of self-training is largely determined by the \textit{out-of-distribution (OOD) generalization} capabilities of the initial teacher model $\theta^{init}_{tch}$, and thus propose a pre-training (``burn-in'') strategy aimed at improving OOD generalization \textit{before} self-training.

We add strong data augmentations including random resizing, color jitter, and cutout~\citep{devries2017improved,chen2020simple}, and keep an EMA copy of the model during burn-in, two strategies that have previously been shown to improve OOD generalization~\citep{morales2024exponential,arpit2022ensemble,gao2022out}, \ie we pre-train a model $\theta$ with the loss from \cref{eq:sup}, where $t \sim T_{\text{src}}$ 
and ${L}_{\text{sup}}$ are still the standard Faster R-CNN losses. 
Each iteration we update an EMA copy of the model, 

\vspace{-7pt}
\begin{equation}
    \theta^{\text{EMA}} = \alpha \theta^{\text{EMA}} + (1 - \alpha) \theta
\end{equation}
\vspace{-11pt}

with \(\alpha \in [0, 1]\). After pre-training, we initialize $\theta_{\text{stu}} = \theta_{\text{tch}} = \theta^{\text{EMA}}$. We are the first to utilize these strategies for DAOD burn-in, and we show in \cref{sec:ablations} that this pre-training strategy leads to faster convergence time and better results. 

\noindent
\textbf{2. Multi-task soft distillation.} Most prior work utilizes confidence thresholding and non-maximum suppression to generate ``hard'' pseudo-labels from teacher predictions $\hat{p}_{tgt}$. However in object detection this strategy is sensitive to the confidence threshold, leading to false positive and false negative errors that harm self-training~\citep{kay2023unsupervised,roychowdhury2019automatic}. Inspired by the knowledge distillation literature we propose instead using ``soft'' distillation losses---\ie using teacher prediction scores as targets without thresholding---allowing us to eliminate the confidence threshold hyperparameter. 

We describe here our approach for two-stage (Faster R-CNN-based) object detection. Distillation implementation details for YOLO and DETR architectures can be found in \cref{sec:implementation}. We distill each task of Faster R-CNN---Region Proposal Network localization ($rpn$) and objectness ($obj$), and Region-of-Interest Heads localization ($roih$) and classification ($cls$)---independently.
At each stage, the teacher provides distillation targets for the same set of input proposals used by the student---\ie anchors $A$ in the first stage, and \textit{student} region proposals $p^{rpn}_{tgt}$ in the second stage:

\begin{minipage}{0.495\linewidth}
    \begin{equation}
        \hspace{-0.32in}
        p^{rpn,obj}_{tgt} = \theta^{rpn,obj}_{stu}(A, x^{t}_{tgt})
    \end{equation}
\end{minipage}
\begin{minipage}{0.495\linewidth}
    \begin{equation}
        \hspace{-0.32in}
        \hat{p}^{rpn,obj}_{tgt} = \theta^{rpn,obj}_{tch}(A, x^{\hat{t}}_{tgt})
    \end{equation}
\end{minipage}
\begin{minipage}{0.495\linewidth}
    \begin{equation}
        p^{roih,cls}_{tgt} = \theta^{roih,cls}_{stu}(p^{rpn}_{tgt}, x^{t}_{tgt})
    \end{equation}
\end{minipage}
\begin{minipage}{0.495\linewidth}
    \begin{equation}
        \hat{p}^{roih,cls}_{tgt} = \theta^{roih,cls}_{tch}(p^{rpn}_{tgt}, x^{\hat{t}}_{tgt})
    \end{equation}
\end{minipage}
\\\\
\noindent
At each iteration, student distillation losses $L_{distill}$ are computed as:

\begin{equation}
L^{rpn}_{distill} = \lambda_{0}L_{rpn}(p^{rpn}_{tgt}, \hat{p}^{rpn}_{tgt}) + \lambda_{1}L_{obj}(p^{obj}_{tgt}, \hat{p}^{obj}_{tgt})
\end{equation}
\begin{equation}
L^{roih}_{distill} = \lambda_{2}L_{roih}(p^{roih}_{tgt}, \hat{p}^{roih}_{tgt}) + \lambda_{3}L_{cls}(p^{cls}_{tgt}, \hat{p}^{cls}_{tgt})
\end{equation}
\begin{equation}
L_{distill} = L^{rpn}_{distill} + L^{roih}_{distill}
\end{equation}

\noindent
Where $L_{rpn}$ and $L_{roih}$ are the smooth L1 loss and $L_{obj}$ and $L_{cls}$ are the cross-entropy loss, and $\lambda_{0\dots3} = 1$ by default. See \cref{fig:framework} for a visual depiction, and the appendix for implementation details. 

One prior DAOD work, PT~\citep{chen2022learning}, has also used soft distillation losses. Our method addresses two shortcomings: (1) PT requires a custom ``Probabilistic R-CNN'' architecture for distillation, while our approach is general and can work with any two-stage detector, and (2) PT uses $\hat{p}^{cls}$ as an indirect proxy for distilling $p^{obj}$, while our approach distills each task directly.

\noindent
\textbf{3. Revisiting DAOD training recipes.} We also re-examine common design choices in DAOD in order to establish strong baseline settings for ALDI++. In particular, we find that two simple changes consistently improve domain adaptation results: (1) Using strong regularization on both target \textit{and} source data during self-training, and (2) Training with equal amounts of source and target supervision in each minibatch (\ie $B_{src} = B_{tgt}$). We also opt to disable all feature alignment in \methodbest to stabilize training and find that the effects on accuracy are minimal (see \cref{sec:ablations}).

\section{The CFC-DAOD Dataset}
\label{sec:data}

\begin{figure}[h]
    \centering
    \begin{minipage}[c]{.48\linewidth}
      \centering
      \includegraphics[width=\linewidth]{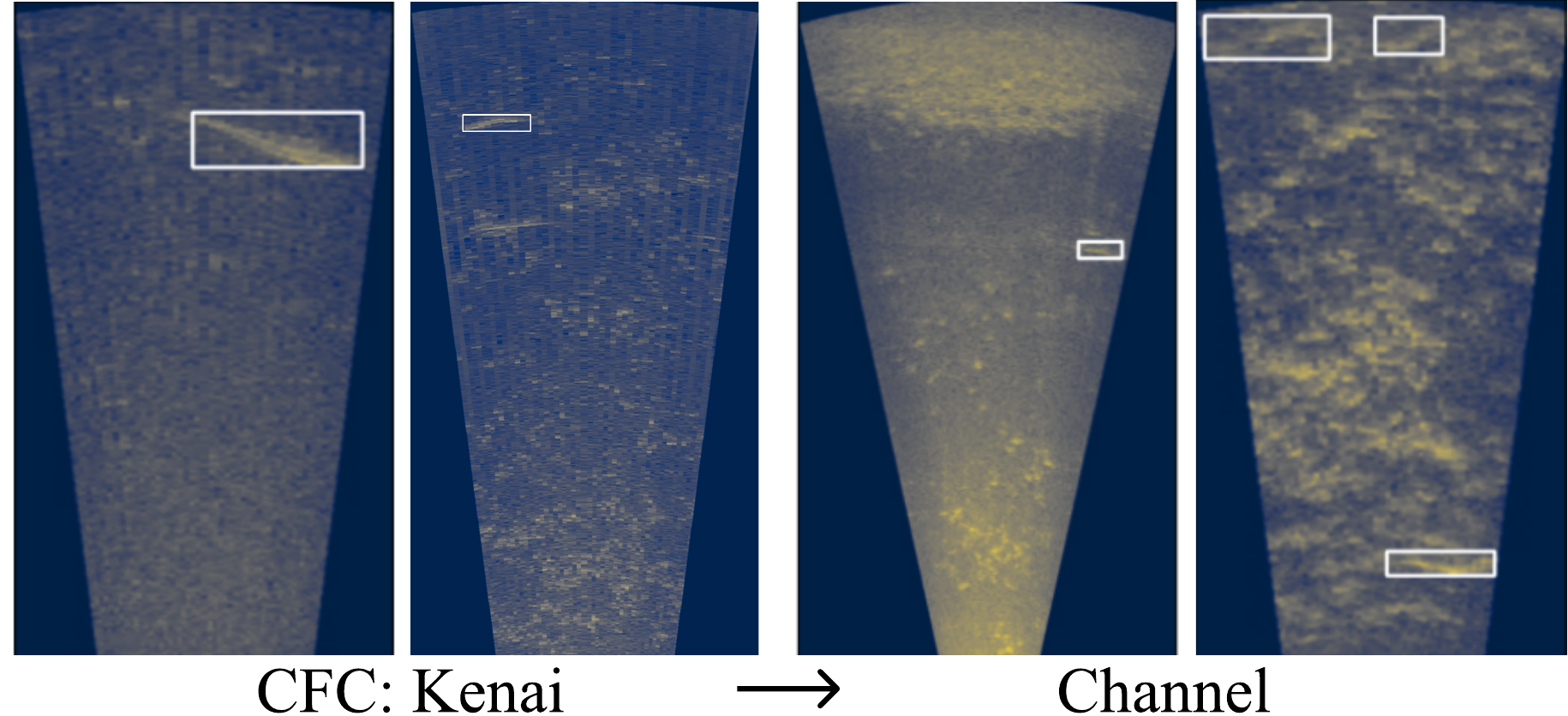}
    \end{minipage}%
    \begin{minipage}[c]{.52\linewidth}
      \centering
      \includegraphics[width=\linewidth]{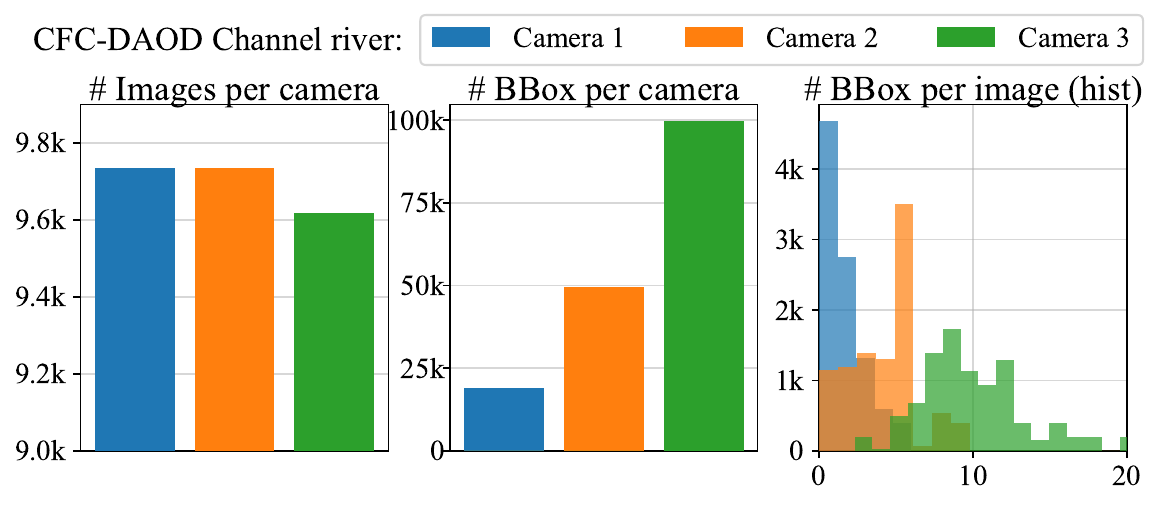} \\
    \end{minipage}
    \caption{\textbf{The CFC-DAOD benchmark} focuses on detecting fish (white bounding boxes) in sonar imagery under domain shift caused by environmental differences between the training location (Kenai) and testing location (Channel). 
    Our dataset contains 168k bounding boxes in 29k frames sampled from 150 new videos captured over two days from 3 different sonar cameras on the Channel river, enabling DAOD experiments. Here we visualize the distribution of images and annotations from each camera.
    }
    \label{fig:cfc}
\end{figure}

Next we introduce our dataset contribution, CFC-DAOD, as a step toward addressing \ul{P3: (a)~Lack of diverse benchmarks leading to overestimation of methods' generality}.

\noindent
\textbf{CFC.} The Caltech Fish Counting Dataset (CFC)~\citep{kay2022caltech} is a \textit{domain generalization} benchmark sourced from fisheries monitoring, where sonar video is used to detect and count migrating salmon. The detection task consists of a single class (``fish'') and domain shift is caused by real-world environmental differences between camera deployments. We identify this application as an opportunity to study the generality of DAOD methods due to its stark differences with existing DAOD benchmarks---specifically, sonar imagery is grayscale, has low signal-to-noise ratios, and foreground objects are difficult to distinguish from the background---however CFC focuses on generalization rather than adaptation and \textit{does not include the data needed for DAOD}.

\noindent
\textbf{CFC-DAOD}
We introduce an extension to CFC, deemed CFC-DAOD, to enable the study of DAOD in this application domain. The task is to adapt from a source location---``Kenai'', \ie the default training set from CFC---to a difficult target location, ``Channel''. 
We collected an additional 168k bounding box annotations in 29k frames sampled from 150 new videos captured over two days from 3 different sensors on the ``Channel'' river (see \cref{fig:cfc}). For consistency, we closely followed the video sampling protocol used to collect the original CFC dataset as described by the authors (see \citep{kay2022caltech}).
Our addition to CFC is crucial for DAOD as it adds an unsupervised training set for domain adaptation methods and a supervised training set to train oracle methods.
We keep the original supervised Kenai training set from CFC (132k annotations in 70k images) and the original Channel test set (42k annotations in 13k images). 
We note this is substantially larger than existing DAOD benchmarks (CS contains 32k instances in 3.5k images, and Sim10k contains 58k instances in 10k images). 
See \cref{appendix:cfc} for more dataset statistics and \cref{sec:qualitative} for qualitative visualizations. We make the dataset public. 

\section{Experiments}
\label{sec:results}

\begin{figure}[h]
    \centering
    \begin{minipage}[c]{.55\linewidth}
      \centering
      \includegraphics[width=\linewidth]{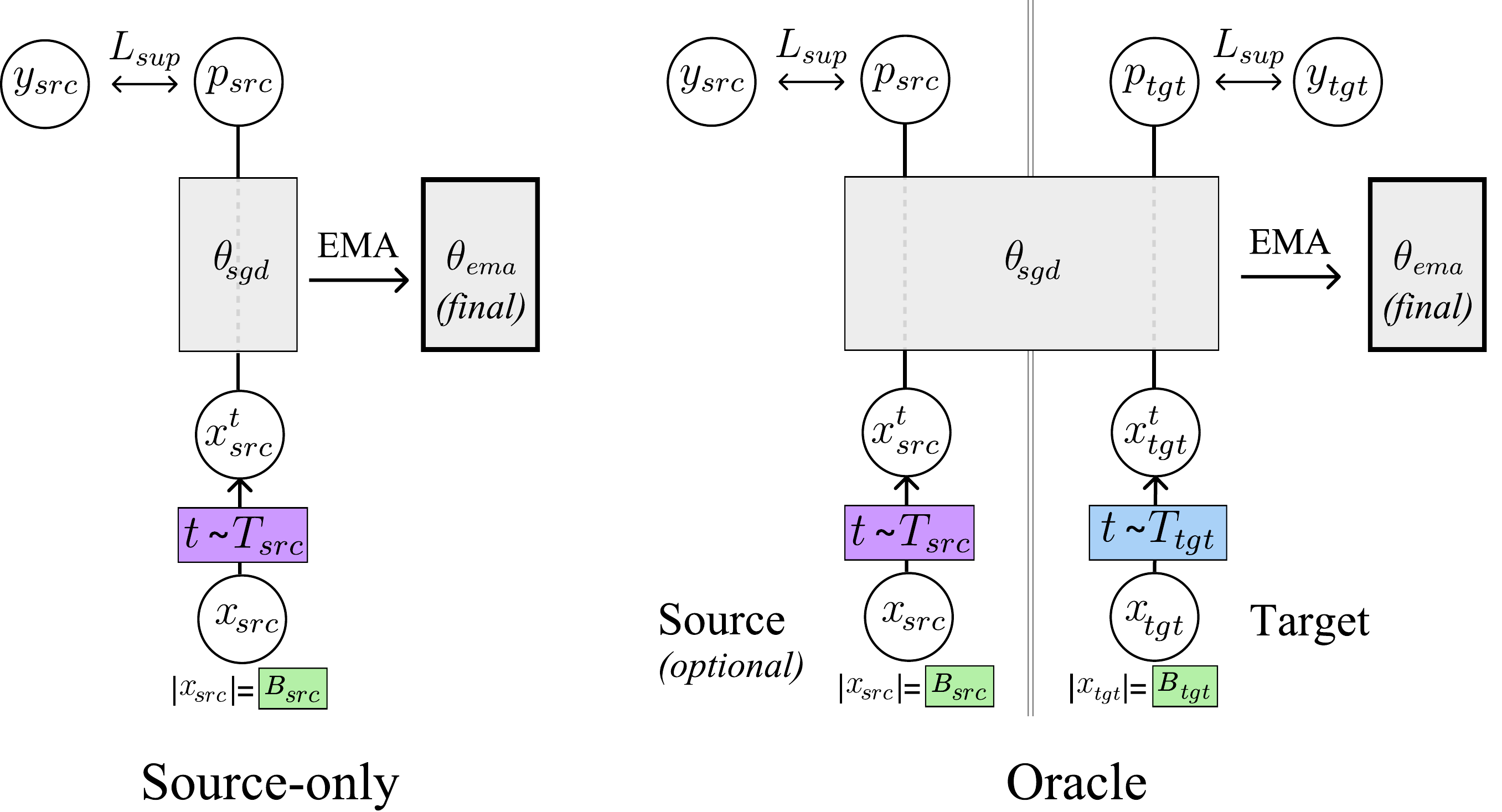}
    \end{minipage}%
    \begin{minipage}[c]{.31\linewidth}
      \centering
      \includegraphics[width=\linewidth]{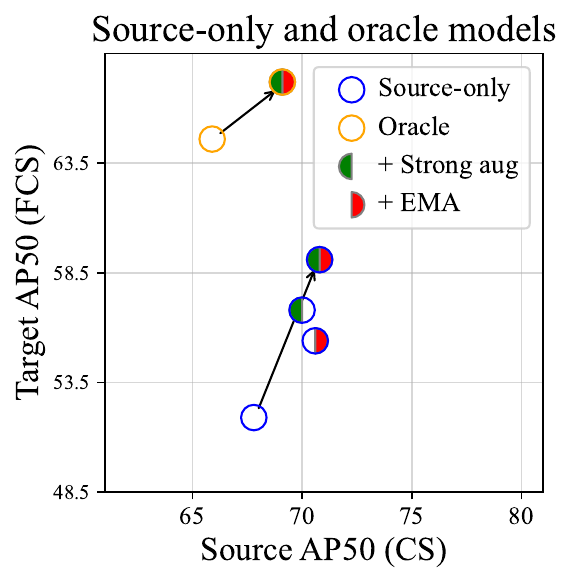} \\
    \end{minipage}
    \caption{\textbf{Revisiting source-only and oracle models in DAOD.} 
    We argue that in order to provide a fair measure of \textit{domain adaptation} performance in DAOD, source-only and oracle models must utilize the same \textit{non-adaptive} architectural and training components as methods being studied. In the case of \textit{Align and Distill}-based approaches, this means source-only and oracle models must have access to the same set of source augmentations and EMA as DAOD methods.
    We see that these upgrades significantly improve source-only performance on target-domain data (+7.2 AP50 on Foggy Cityscapes), \textit{even though the source-only model has never seen any target-domain data}, and these upgrades also improve oracle performance. Overall, these results set more challenging and realistic performance targets for DAOD methods.
    }
    \label{fig:baseline_oracle}
\end{figure}

\noindent
In this section we propose an updated benchmarking protocol for DAOD (\cref{sec:baselinesoracles}) that allows us to fairly analyze the performance of \methodbest compared to prior work (\cref{sec:final_results}) and conduct extensive ablation studies (\cref{sec:ablations}).

\subsection{Benchmarking Protocol}
\label{sec:baselinesoracles}

\noindent
\textbf{Datasets.} We perform experiments on Cityscapes $\rightarrow$ Foggy Cityscapes, Sim10k $\rightarrow$ Cityscapes, and CFC Kenai $\rightarrow$ Channel. In addition to being consistent with prior work, these datasets represent three common adaptation scenarios capturing a range of real-world challenges: weather adaptation, Sim2Real, and environmental adaptation, respectively. We note that there have been inconsistencies in prior work in terms of which ground truth labels for Cityscapes are used. 
We use the Detectron2 version, which includes three intensity levels of fog $\{0.005, 0.01, 0.02\}$.

\noindent
\textbf{Metrics.} For all experiments we report the PascalVOC metric of mean Average Precision with IoU $\geq 0.5$ (``AP50'')~\citep{everingham2010pascal}. This is consistent with prior work on Cityscapes, Foggy Cityscapes, Sim10k, and CFC.

\noindent
\textbf{Revisiting source-only and oracle models.} Here we address \ul{P1: Improperly constructed source-only and oracle models, leading to overestimation of performance gains}. The goal of DAOD is to develop adaptation techniques that use unlabeled target-domain data to improve target-domain performance. 
Thus, in order to properly isolate \textit{adaptation-specific} techniques, \textbf{any technique that does not need target-domain data to run should also be used by source-only and oracle models}.
This means that source-only and oracle models should also utilize the same strong augmentations and EMA updates as DAOD methods.

In \cref{fig:baseline_oracle} we illustrate the resulting source-only and oracle models, and show that including these components significantly improves both source-only and oracle model performance (+7.2 and +2.6 AP50 on Foggy
Cityscapes, respectively). This has significant implications for DAOD research: 
because source-only and oracle models have not been constructed with equivalent components, performance gains stemming from better generalization have until now been \textit{misattributed} to DAOD. With properly constructed source-only and oracle models, the gains from DAOD are much more modest (see \cref{fig:sota}). Note that for clarity we compare all methods in \cref{fig:fig1} and \cref{fig:sota} against a single source-only and oracle model, however it would be more appropriate to compare each method to its own bespoke source-only and oracle models that use the same training components; see \cref{sec:method-specific} for the full comparison.

\noindent
\textbf{Fixed training settings.} Prior work has used inconsistent backbones and image sizes, making head-to-head comparisons less fair. Using ALDI we instead compare using the same training settings, offering a fair comparison. As a starting point we utilize reasonably modern settings likely to be used by a practitioner: the Cityscapes defaults in the Detectron2 codebase. All methods in our comparisons, including source-only and oracle models, utilize Faster R-CNN architectures with ResNet-50~\citep{ren2015faster} backbones with FPN~\citep{lin2017feature}, COCO~\citep{lin2014microsoft} pre-training, and an image size of 1024px on the shortest side. See \cref{appendix:experiment_settings} for more details.

\subsection{Fair Comparison and State-of-the-Art Results}
\label{sec:fair_comparison}
\label{sec:final_results}

\begin{figure}[h]
    \centering    
    \includegraphics[width=\linewidth]{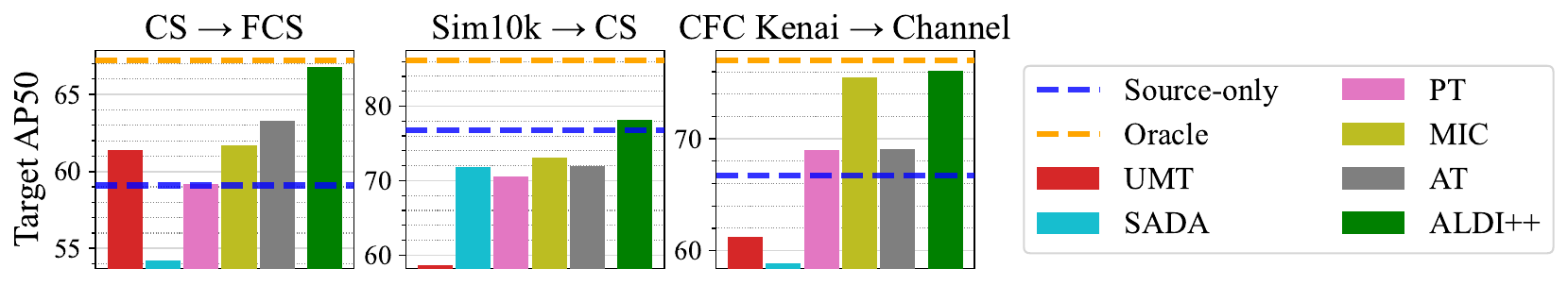}
    \vspace{-12pt}
    \caption{
    \textbf{Fair comparison of \methodbest with  existing state-of-the-art approaches using the \methodname framework and modern training recipes.}
    Some prior methods show consistent benefit but others lag behind fair source-only models. Our method \methodbest outperforms prior work on all datasets studied by a significant margin: +3.5 AP50 on CS $\rightarrow$ FCS, +5.7 AP50 on Sim10k $\rightarrow$ CS, and +0.6 AP50 on CFC Kenai $\rightarrow$ Channel. Notably, \methodbest is the \textit{only} method to outperform a source-only model on Sim10k $\rightarrow$ CS.
    }
    \label{fig:sota}
\end{figure}

\noindent
We compare \methodbest with reimplementations of five state-of-the-art DAOD methods on top of our framework: UMT~\citep{deng2021unbiased}, SADA~\citep{chen2021scale}, 
PT~\citep{chen2022learning}, MIC~\citep{hoyer2023mic}, and AT~\citep{li2022cross}; see \cref{appendix:experiment_settings} for the \methodname settings used to reproduce them. We use the fair benchmarking protocol proposed in \cref{sec:baselinesoracles}. 
Results are shown in \cref{fig:sota}. All methods (including \methodbest) use the same settings for all benchmarks.

\noindent
\textbf{Comparison with state-of-the-art.} \methodbest outperforms all prior work and sets a new state-of-the-art on all benchmarks studied, outperforming the next-best methods by +3.5 AP50 on CS $\rightarrow$ FCS, +5.7 AP50 on Sim10k $\rightarrow$ CS (where ours is the \textit{only} method to outperform a fair source-only model), and +0.6 AP50 on CFC Kenai $\rightarrow$ Channel.
ALDI++ achieves near-oracle level performance on CS $\rightarrow$ FCS and CFC Kenai $\rightarrow$ Channel (0.4 and 0.9 AP50 away, respectively), while other methods close less than half the gap between source-only and oracle models.

\noindent
\textbf{Comparison across datasets.} We compare all methods on CS $\rightarrow$ FCS, Sim10k $\rightarrow$ CS, and CFC Kenai $\rightarrow$ Channel, in \cref{fig:sota}. We find the ranking of methods differs across datasets. ALDI++, MIC and AT are consistently the top-performing methods across all datasets. MIC performs noticeably better on CFC Kenai $\rightarrow$ Channel than other prior work, nearly matching the performance of ALDI++. UMT exhibits variable performance due to the differences in the difficulty of image generation across datasets (see \cref{appendix:experiment_settings} for examples). SADA underperforms other methods on CS $\rightarrow$ FCS and CFC Kenai $\rightarrow$ Channel, but closes this gap on the more difficult Sim10k $\rightarrow$ CS. These results demonstrate the utility of CFC-DAOD as another point of comparison for DAOD methods; we see that method performance on synthetic benchmarks like CS $\rightarrow$ FCS is not necessarily indicative of performance on real-world domain shifts.

\noindent
\textbf{Comparison with fair source-only and oracle models}. Re-implementing methods in ALDI improves absolute performance of most methods due to upgraded training settings; however performance decreases dramatically compared to source-only and oracle models. There are several instances where modernized DAOD methods are actually \textit{worse} than a fair source-only model. 
Notably, a source-only model outperforms upgraded versions \textit{all} previously-published work on Sim10 $\rightarrow$ CS. 
We also see that no state-of-the-art methods outperform a fair oracle on any dataset, in contrast to claims made by prior work~\citep{li2022cross,chen2022learning,cao2023contrastive}.

\begin{figure}
    \centering
    \includegraphics[width=0.85\linewidth]{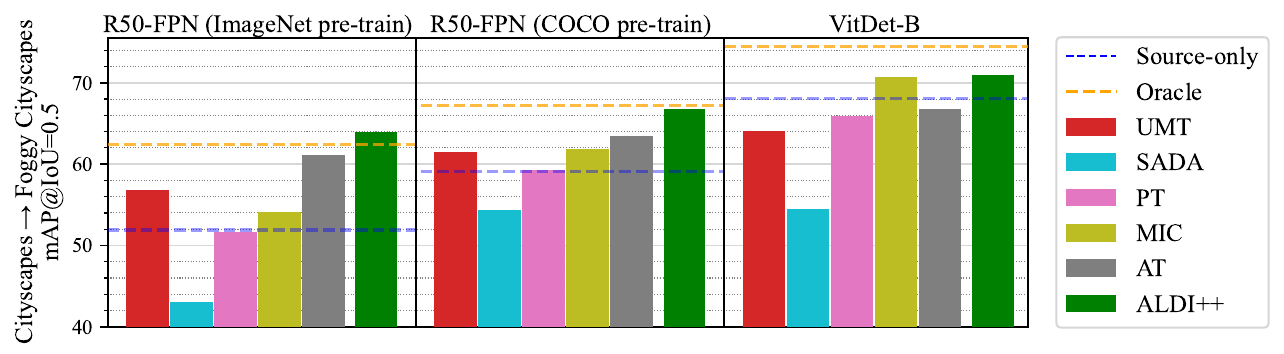}
    \vspace{-5pt}
    \caption{\textbf{Comparison across backbones and pre-training data.} Pre-training strategy does not significantly change the relative strength of methods compared to each other nor compared to source-only and oracle models. However, all models perform worse compared to source-only and oracle models when using VitDet backbones, with only ALDI++ and MIC outperforming a source-only model.}
    \vspace{-9pt}
    \label{fig:pretraining}
\end{figure}

\textbf{Comparison across backbones and pre-training data.} We compare all methods on CS $\rightarrow$ FCS using ImageNet vs. COCO pre-training, as well as Resnet-50 FPN backbones vs. VitDet-B backbones, in \cref{fig:pretraining}. See \cref{appendix:tables} for other datasets. We see that while COCO pre-training improves absolute performance of all methods, their ranking does not change significantly compared to ImageNet pre-training, nor does their performance in relation to source-only baselines. Interestingly, we find that ALDI++ outperforms an oracle on CS $\rightarrow$ FCS when using ImageNet pre-training, however we do not observe this trend on the other datasets (see \cref{appendix:tables}). We hypothesize that this may be due to noise in the Foggy Cityscapes target-train labels due to their programmatic generation, and that COCO pre-training helps prevent the oracle from overfitting to these erroneous boxes. We show that only ALDI++ and MIC continue to show improvements over an upgraded VitDet source-only model (see \cref{appendix:vit} for other datasets), with ALDI++ performing slightly better than MIC (+0.4 AP50). 
We see there is a larger gap between the ViT \methodbest and the ViT oracle compared to ResNet backbones, indicating the potential for future work to improve performance. Across all experiments in \cref{fig:pretraining} we see that the source-only--oracle gap shrinks as the underlying model improves due to either stronger pre-training or backbone upgrades, indicating that DAOD may offer diminishing returns with stronger models.

\subsection{Ablation Studies}
\label{sec:ablations}

In this section we ablate the performance of each component of \methodname on CS~$\rightarrow$~FCS. 

\noindent
\textbf{Base settings.} For each ablation, unless otherwise specified we begin with the following training settings. We initialize {$\theta_{stu}, \theta_{tch}$} with COCO pre-training followed by a burn-in phase on $X_{\text{src}}$ with weak augmentations and early stopping based on validation performance. $T_{\text{src}}$ includes random horizontal flip and random scaling. $T_{\text{tgt}}$ includes random horizontal flip, random scaling, color jitter, and cutout. The $B_{\text{src}}:B_{\text{tgt}}$ batch ratio is 1:1. $L_{\text{distill}}$ is hard pseudo-labeling with a confidence threshold of 0.8, and $L_{\text{align}}$ is disabled. Note these base settings are not necessarily those of \methodbest but rather the most commonly chosen values in prior work for each component. Additional training settings are reported in \cref{appendix:experiment_settings}.

\begin{figure}[!b]
    \centering
    \includegraphics[width=\linewidth]{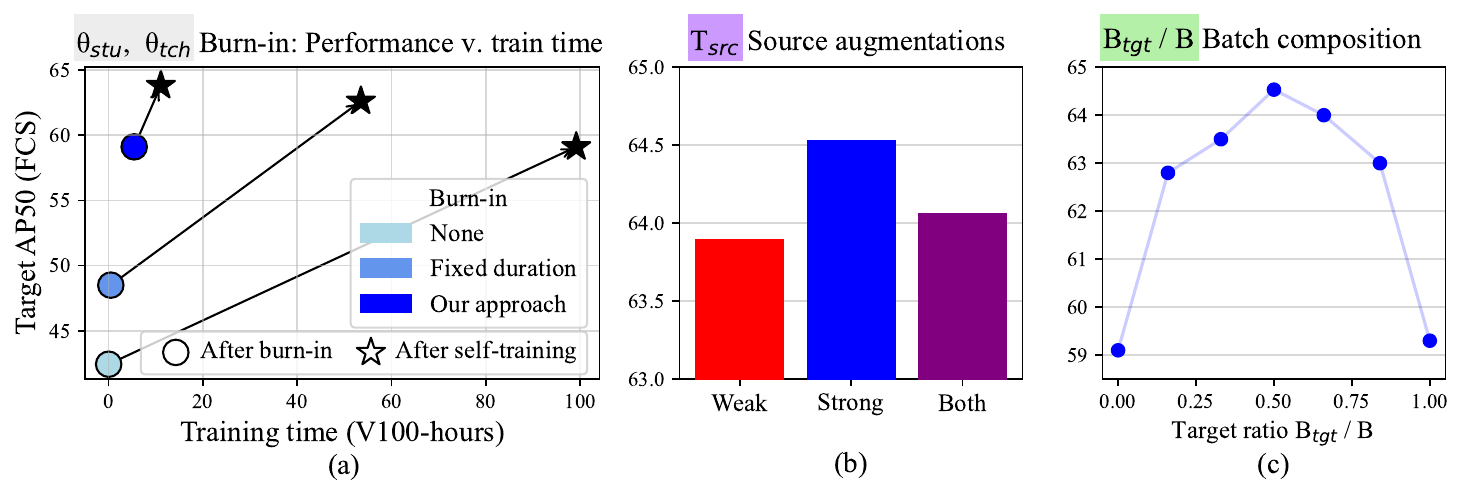}
    \vspace{-20pt}
    \caption{
    \textbf{Ablation studies.} 
    \textbf{(a)}~Our proposed burn-in strategy (\cref{sec:ours}) improves AP50$_{FCS}$ by +4.7 and reduces training time by 10x compared to no burn-in. \textbf{(b)}~Strong source-data augmentations during self-training lead to better performance. \textbf{(c)}~An equal ratio of source and target data during self-training leads to best performance.
    }
    \label{fig:ablations}
\end{figure}

\noindent
\textbf{\colorbox{Burnin}{$\theta_{stu}$, $\theta_{tch}$} Network initialization (burn-in).}
In \cref{fig:ablations}a we analyze the effects of our proposed burn-in strategy (see \cref{sec:ours}). 
We measure performance in terms of target-domain AP50 as well as convergence time, defined as the training time at which the model first exceeds 95\% of its final target-domain performance. 
We compare our approach with: (1) No dataset-specific burn-in, \ie starting with COCO weights, and (2) The approach used by past work---using a fixed burn-in duration, \eg 10k iterations. 
We find that our method results in significant improvements in both training speed and accuracy, 
leading to upwards of 10\% improvements in AP50 and reducing training time by a factor of 10 compared to training without burn-in.

\noindent
\textbf{\colorbox{T_src}{$T_{src}$} Source augmentations.} In \cref{fig:ablations}b we ablate the set of source-domain data augmentations.
We compare using weak augmentations (random flipping and random scaling), strong augmentations (color jitter and cutout), and a combination of weak and strong, noting that prior works differ in this regard but do not typically report the settings used. We find that using strong source augmentations on the entire source-domain training batch outperforms weak augmentations and a combination of both.

\noindent
\textbf{\colorbox{T_tgt}{$T_{tgt}$} Target augmentations.} In \cref{tab:target_aug} we investigate the use of different augmentations for target-domain inputs to the \textit{student} model. (We note that weak augmentations are always used for target-domain inputs to the teacher in accordance with prior work). We see that stronger augmentations consistently improve performance, with best performance coming from the recently-proposed MIC augmentation~\citep{hoyer2023mic}.

\begin{table}[t]
\caption{\textbf{Ablation studies.} 
\textbf{(a)}~Effects of target-domain augmentation on self-training. Augmentations applied to student inputs ($T_{tgt}$ in \cref{fig:framework}). Stronger augmentations improve performance considerably. \textbf{(b)}~Effects of distillation objectives on self-training. We compare hard targets---used by most prior art, which thresholds teacher predictions to create pseudo-labels---with our proposed soft targets. Soft targets can improve overall performance. Results are the mean and standard deviation over 3 runs. \textbf{(c)}~Feature alignment has diminishing returns. Compared to a source-only baseline AP50 of 59.1, feature alignment objectives $L_{align}$ without self-training provides up to 2.6 AP50 of benefit (first row), but diminishes to 0.2 AP50 additional gain when used alongside self-training (last row).}
\vspace{-5pt}

\resizebox{\linewidth}{!}{
\hspace{15pt}
\subfloat[\label{tab:target_aug}]{
    \centering
    \begin{tabular}{l|r}
         \cellcolor{T_tgt}{{$T_{tgt}$}} & AP50$_{FCS}$  \\
         \hline\hline
         
         
         \rule{0pt}{\normalbaselineskip}Weak (scale \& flip) & 52.6 \\
         + Color jitter & 59.0 \\
         + Color jitter + Erase & 63.1 \\
         + Color jitter + MIC & 64.3 
    \end{tabular}
    }
    \hspace{15pt}
    \subfloat[\label{tab:distill_ablate}]{
    \centering
    \begin{tabular}{l|l}
        \cellcolor{L_distill}{{$L_{distill}$}} & AP50$_{FCS}$ \\
        \hline\hline
        
        
        \rule{0pt}{\normalbaselineskip}Hard targets & 63.7 $\pm$ 0.1 \\
        Soft targets & 64.0 $\pm$ 0.4
    \end{tabular}
    }
    \centering
    \subfloat[\label{tab:align}]{
    \centering
    \hspace{15pt}
    \begin{tabular}{c|c|l}
        \cellcolor{L_align}{{$L_{align}$}} & \cellcolor{L_distill}{{$L_{distill}$}} & AP50$_{FCS}$ \\
        \hline\hline
        
        
        \rule{0pt}{\normalbaselineskip}\checkmark & & 61.7 \\
        \rule{0pt}{\normalbaselineskip} & \checkmark & 63.7 \\
        \rule{0pt}{\normalbaselineskip}\checkmark & \checkmark & 63.9 \\
    \end{tabular}
    
    }
    \hspace{15pt}
}
\vspace{-10pt}
\end{table}

\noindent
\textbf{\colorbox{MtM}{$B_{tgt}/B$} Batch composition.} In \cref{fig:ablations}c we ablate the ratio of source and target data within a minibatch. We note that prior works differ in this setting but do not typically report what ratio is used. We see that using equal amounts of source and target data within each minibatch leads to the best performance. Notably, we also find that the inclusion of source-domain imagery is essential to see benefits from self-training---without any source imagery, AP50$_{FCS}$ drops from 64.5 to 59.3.

\noindent
\textbf{\colorbox{L_distill}{$L_{distill}$} Self-distillation.}
In \cref{tab:distill_ablate} we analyze the effects of our proposed multi-task soft distillation approach (see \cref{sec:ours}). 
We compare our approach with the ``hard'' pseudo-label approach used by prior work,
where teacher predictions are post-processed  with non-maximum suppression and a hard confidence threshold of 0.8 \citep{hoyer2023mic,li2022cross,deng2021unbiased,liu2021unbiased}. 
For our proposed ``soft'' distillation method, we first sharpen teacher predictions at both detector stages using a sigmoid for objectness predictions and a softmax for classification predictions, both with a default temperature of 1. 
We see that our proposed soft targets improve performance compared to hard targets.

\noindent
\textbf{\colorbox{L_align}{$L_{align}$} Feature alignment.}
Finally we investigate the use of feature alignment. 
We implement an adversarial feature alignment approach consisting of an image-level and instance-level feature discriminator (our implementation performs on par with SADA while being simpler to train; see \cref{appendix:feature_align}). In \cref{tab:align}, we show that feature alignment used in isolation (\ie without self-training) offers performance gains up to 2.6 AP50. However, these performance gains are smaller than those seen from self-training (AP50$_{FCS}$ of 61.7 vs. 63.7, respectively). When used in combination with self-training techniques, the additional benefit of feature alignment drops to $\leq$~0.2 AP50$_{FCS}$. This suggests that self-training is currently the most promising avenue for progress and that more research is needed to develop complementary approaches. We also note that feature alignment approaches introduce training instability that may not be worth the small performance gain for practical use.

\vspace{-5pt}
\section{Discussion and Conclusions}
\label{sec:conclusion}

In this work we proposed: the ALDI framework and an improved DAOD benchmarking methodology, providing a critical reset for the DAOD research community; a new dataset CFC-DAOD, increasing the diversity and real-world applicability of DAOD benchmarks; and a new method ALDI++ that advances the state-of-the-art.
We conclude with key findings.

\noindent
\textbf{Network initialization has an outsized impact.} We find that general advancements in computer vision eclipse progress in DAOD: a Resnet50-FPN source-only model outperforms all VGG-based DAOD methods, and a VitDet source-only model outperforms all Resnet50-FPN based DAOD methods.
Similarly, simply adding stronger augmentations and EMA to source-only models leads to \textit{better target-domain performance than some adaptation methods}, and including these upgrades during network initialization (burn-in) improves adaptation performance as well. 

\noindent
\textbf{DAOD techniques are helpful, but do not consistently achieve oracle-level performance as previously claimed~\citep{li2022cross,chen2022learning,cao2023contrastive}.}
Top-performing DAOD methods, including \methodbest, demonstrate improvements over source-only models (see \cref{fig:fig1} and \cref{fig:sota}). However, in contrast to previously-published results, no DAOD method consistently reaches oracle-level performance across datasets, architectures, and pre-training strategies, suggesting there is still room for improvement. The gap between DAOD methods and oracles is even larger for stronger architectures like VitDet. This is a promising area for future research.

\noindent
\textbf{Benchmarks sourced from real-world domain adaptation challenges can help the community develop generally useful methods.} 
We find that DAOD methods do not necessarily perform equivalently across datasets (see \cref{fig:sota}). Diverse benchmarks are useful to make sure we are not overfitting to the challenges of one particular use case, while exposing and supporting progress in impactful applications. 
Our contributed codebase and benchmark dataset provide the necessary starting point to enable this effort.

\noindent
\textbf{A lack of transparent comparisons has incentivized incremental progress in DAOD.} 
Most highly-performant prior works in DAOD are some combination of DANN~\citep{ganin2016domain} and Mean Teacher~\citep{tarvainen2017mean} plus custom training techniques. Without fair comparisons it has been possible to propose near-duplicate methods that still achieve state-of-the-art performance due to hyperparameter tweaks.
Our method \methodbest establishes a strong point of comparison for \textit{Align and Distill}-based approaches that will require algorithmic innovation to surpass.

\noindent
\textbf{Validation is the elephant in the room.} All of our experiments, and all previously published work in DAOD, utilize a target-domain validation set to perform model and hyperparameter selection. This violates a key assumption in unsupervised domain adaptation: that no target-domain labels are available to begin with. Prior work has shown that it may not be possible to achieve performance improvements in domain adaptation \textit{at all} under realistic validation conditions~\citep{musgrave2021unsupervised,musgrave2022benchmarking,kay2023unsupervised}. Therefore our results (as well as previously-published work) can really only be seen as an upper bound on DAOD performance. While this is valuable, further research is needed to develop effective unsupervised validation procedures for DAOD.

\section*{Acknowledgments}

This material is based upon work supported by: NSF CISE Graduate Fellowships Grant \#2313998, MIT EECS department fellowship \#4000184939, MIT J-WAFS seed grant \#2040131, Caltech Resnick Sustainability Institute Impact Grant ``Continuous, accurate and cost-effective counting of migrating salmon for conservation and fishery management in the Pacific Northwest'', NSF Award \#2330423 and NSERC Award \#585136. Any opinions, findings, and conclusions or recommendations expressed in this material are those of the authors and do not necessarily reflect the views of NSF, NSERC, MIT, J-WAFS, Caltech, or RSI. The authors acknowledge the MIT SuperCloud and Lincoln Laboratory Supercomputing Center for providing HPC resources~\cite{reuther2018interactive}. We also thank the Alaska Department of Fish and Game for their ongoing collaboration and for providing data, and Sam Heinrich, Neha Hulkund, Kai Van Brunt, Rangel Daroya, and Mark Hamilton for helpful feedback.

\bibliography{main}
\bibliographystyle{tmlr}

\clearpage

\appendix

\section{Additional Experiments}

\subsection{YOLO and DETR Architectures}
\label{appendix:yolodetr}

To demonstrate the architecture-agnosticism of our framework and enable further research, we implement ALDI for the one-stage detection architecture YOLOv5~\cite{redmon2016lookonceunifiedrealtime,yolov8_ultralytics} and the transformer-based architecture Deformable DETR~\cite{carion2020end,zhu2020deformable}. Implementation details are further described in \cref{sec:implementation}.

\begin{table}[h]
    \centering
    \caption{\textbf{ALDI-YOLO results.}}

\resizebox{\linewidth}{!}{

    \subfloat[CS $\rightarrow$ FCS]{
    \centering
    \begin{tabular}{c|r}
         Method &  AP50\\
         \hline\hline
         \rule{0pt}{\normalbaselineskip}{Source-only (YOLOv5m)} & {58.8} \\
         \hline
         \rule{0pt}{\normalbaselineskip}SSDA-YOLO~\cite{zhou2023ssda} (YOLOv5l) & 55.9 \\
         \textbf{ALDI-YOLO (ours, YOLOv5m)} & \textbf{62.5} \\
         \hline
         \rule{0pt}{\normalbaselineskip}{Oracle (YOLOv5m)} & {66.3} \\
    \end{tabular}
    }

    \subfloat[Sim10k $\rightarrow$ CS]{
    \centering
    \begin{tabular}{c|r}
         Method &  AP50\\
         \hline\hline
         \rule{0pt}{\normalbaselineskip}{Source-only} & {75.0} \\
         \hline
         \rule{0pt}{\normalbaselineskip}ALDI-YOLO (ours) & 75.0\\
         \hline
         \rule{0pt}{\normalbaselineskip}Oracle & 88.0
    \end{tabular}
    }

    \subfloat[CFC-DAOD]{
    \centering
    \begin{tabular}{c|r}
         Method &  AP50\\
         \hline\hline
         \rule{0pt}{\normalbaselineskip}{Source-only} & 60.2 \\
         \hline
         \rule{0pt}{\normalbaselineskip}ALDI-YOLO (ours) & 52.4\\
         \hline
         \rule{0pt}{\normalbaselineskip}Oracle & 76.7
    \end{tabular}
    }
    
}
\end{table}

\begin{table}[h]
    \centering
    \caption{\textbf{DETR results on Cityscapes $\rightarrow$ Foggy Cityscapes.} We use 800px input size for consistency with prior work.}
    \begin{tabular}{c|r}
         Method &  AP50\\
         \hline\hline
         \rule{0pt}{\normalbaselineskip}{Source-only} & 44.5 \\
         \hline
         \rule{0pt}{\normalbaselineskip}{SFA~\cite{wang2021exploring}} & 41.3 \\
         MTTrans~\cite{yu2022mttrans} & 43.4 \\
         PM-DETR~\cite{jia2023pm} & 44.3 \\
         \textbf{ALDI-DETR (ours)} & \textbf{44.8} \\
         \hline
         \rule{0pt}{\normalbaselineskip}{Oracle} & 50.0 \\
    \end{tabular}
    \label{tab:detr}
\end{table}

\subsection{ViT and ConvNeXt backbones}
\label{appendix:vit}

\begin{table}[h]
\centering
\begin{minipage}{0.4\linewidth}
    \centering
    \caption{Sim10k $\rightarrow$ Cityscapes}
    \begin{tabular}{c|l}
         Method &  AP50$_{CS}$\\
         \hline\hline
         \rule{0pt}{\normalbaselineskip}{ViT-B baseline} & 81.7 \\
         ALDI++ + ViT-B & 81.8\\
         {ViT-B oracle} & 89.8 \\
    \end{tabular}
    \label{tab:vit1}
\end{minipage}%
\hspace{0.04\linewidth}
\begin{minipage}{0.4\linewidth}
    \centering
    \caption{CFC Kenai $\rightarrow$ Channel}
    \begin{tabular}{c|l}
         Method &  AP50$_{Channel}$\\
         \hline\hline
         \rule{0pt}{\normalbaselineskip}{ViT-B baseline} & 69.0 \\
         ALDI++ + ViT-B & 71.1\\
         {ViT-B oracle} & 76.7 \\
    \end{tabular}
    \label{tab:vit2}
\end{minipage}
\end{table}

\begin{table}[h]
\centering
\begin{minipage}{0.4\linewidth}
    \centering
    \caption{Cityscapes $\rightarrow$ Foggy Cityscapes}
    \begin{tabular}{c|l}
         Method &  AP50$_{FCS}$\\
         \hline\hline
         \rule{0pt}{\normalbaselineskip}{ViT-L baseline} & 70.2 \\
         ALDI++ + ViT-L & 76.1\\
         {ViT-L oracle} & 77.4 \\
    \end{tabular}
    \label{tab:vitl}
\end{minipage}%
\hspace{0.04\linewidth}
\begin{minipage}{0.4\linewidth}
    \centering
    \caption{Cityscapes $\rightarrow$ Foggy Cityscapes}
    \begin{tabular}{c|l}
         Method &  AP50$_{FCS}$\\
         \hline\hline
         \rule{0pt}{\normalbaselineskip}{ConvNext-L baseline} & 58.9 \\
         ALDI++ + ConvNext-L & 63.3\\
         {ConvNext-L oracle} & 64.1 \\
    \end{tabular}
    \label{tab:convnext}
\end{minipage}
\end{table}

For completeness we show results using ALDI++ in combination with VitDet{-B}~\cite{li2022exploring} in \cref{tab:vit1} (Sim10k $\rightarrow$ Cityscapes) and \cref{tab:vit2} (CFC Kenai $\rightarrow$ Channel). We see that ALDI continues to demonstrate improvements over baselines even as overall architectures get stronger, though these improvements are smaller in magnitude than VitDet-B results on the CS $\rightarrow$ FCS dataset.

We also demonstrate the performance of ALDI++ with even larger backbones to examine how performance and domain gaps change. We show results from VitDet-L in \cref{tab:vitl}, and with ConvNeXt-L in \cref{tab:convnext}. Results are similar to our main results; we continue to see improvements of baselines, oracles, and ALDI++ in these settings.

\subsection{Adversarial Feature Alignment}
\label{appendix:feature_align}

We report additional ablations for the adversarial feature alignment network(s) used, comparing our implementations of image-level alignment and instance-level alignment with a baseline and SADA.
As we see in \cref{tab:alignment}, \cref{tab:alignment_sim10k}, and \cref{tab:alignment_cfc}, the best settings to use differ by dataset. 
By default our feature alignment experiments in Sec. 6.1 of the main paper use both instance and image level alignment. See \cref{sec:alignment_implementation} below for further implementation details.

\begin{table}[h]
    \centering
    \caption{\textbf{Comparison of adversarial alignment methods.} \textbf{(a)} Cityscapes $\rightarrow$ Foggy Cityscapes. We see that our implementations outperform SADA~\cite{chen2021scale} while being simpler. Image-level alignment is best, followed by Image + Instance.
    \textbf{(b)} Sim10k $\rightarrow$ Cityscapes. Instance-level alignment is best.
    \textbf{(c)} CFC Kenai $\rightarrow$ Channel. Image + Instance is best. We see there is no consistently-best strategy across datasets; however, we note that for all datasets, the benefit of using adversarial feature alignment is smaller than self-training (see Sec. 6.3 of the main paper).
    }

\resizebox{\linewidth}{!}{

    \subfloat[\label{tab:alignment}]{
    \centering
    \begin{tabular}{c|l}
         Method &  AP50$_{FCS}$\\
         \hline\hline
         \rule{0pt}{\normalbaselineskip}{Source-only} & {51.9} \\
         \hline
         \rule{0pt}{\normalbaselineskip}SADA & 54.2 \\
         Image-level (ours) & 55.8\\
         Instance-level (ours) & 54.3 \\
         Image + Instance (ours) & 54.9
    \end{tabular}
    }

    \subfloat[\label{tab:alignment_sim10k}]{
    \centering
    \begin{tabular}{c|l}
         Method &  AP50$_{CS}$\\
         \hline\hline
         \rule{0pt}{\normalbaselineskip}{Source-only} & {70.8} \\
         \hline
         \rule{0pt}{\normalbaselineskip}Image-level (ours) & 71.8\\
         Instance-level (ours) & 73.3 \\
         Image + Instance (ours) & 71.5
    \end{tabular}
    }

    \subfloat[\label{tab:alignment_cfc}]{
    \centering
    \begin{tabular}{c|l}
         Method &  AP50$_{Channel}$\\
         \hline\hline
         \rule{0pt}{\normalbaselineskip}{Source-only} & 65.8 \\
         \hline
         \rule{0pt}{\normalbaselineskip}Image-level (ours) & 65.2\\
         Instance-level (ours) & 66.0 \\
         Image + Instance (ours) & 66.9
    \end{tabular}
    }
    
}
\end{table}

\subsection{Visualizing Alignment}

\begin{figure*}[t]
    \centering
    \includegraphics[width=1\linewidth]{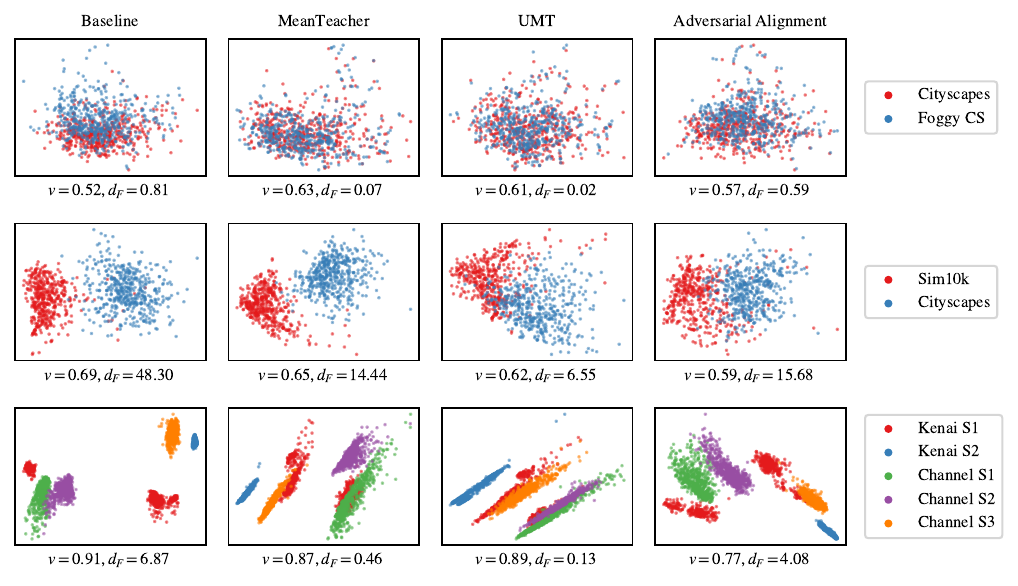}
    \caption{Embedding of pooled features from the final backbone layer in 2D space using PCA. We compare: (1) A source-only baseline, (2) The base settings from \cref{sec:ablations} (``MeanTeacher''), (3) UMT (which utilizes image-to-image translation), and (4) MeanTeacher + adversarial feature alignment using image-level features. The ratio of variance explained by the first two PCA components is given by $v$ and a dissimilarity score between source and target features is given by $d_F$. $d_F$ is lower than the baseline for all alignment methods and does roughly match the overall visual trend in feature overlap. 
    In all cases, the simple MeanTeacher model significantly reduces the distance between source and target data even though there is no explicit alignment criterion, even resulting in a smaller $d_F$ than adversarial alignment methods for CS $\rightarrow$ FCS \& CFC Kenai $\rightarrow$ Channel. 
    }
    \label{fig:feature_vis_img}
\end{figure*}

We investigate the overlap of source and target data in the feature space of different methods. For each method, we pool the highest-level feature maps of the backbone, either globally (``image-level'') or per instance (``instance-level''). We then embed the pooled feature vectors in 2D space using PCA for visual inspection (see Fig~\ref{fig:feature_vis_img}).
We also compute a dissimilarity score based on FID \cite{heusel2017gans}, by fitting Gaussians to the source and target features and then computing the Fréchet distance between them.

\subsection{Teacher update}
\label{appendix:teacher}

\begin{table}
    \centering
    \caption{Comparison of teacher update approaches on Cityscapes $\rightarrow$ Foggy Cityscapes. Mean teacher greatly outperforms other options.
    }
    \begin{tabular}{c|l}
         Method &  AP50$_{FCS}$\\
         \hline\hline
         \rule{0pt}{\normalbaselineskip}{Source-only baseline} & 51.9 \\
         \hline
         \rule{0pt}{\normalbaselineskip}No update (vanilla self-training) & 52.9\\
         Student is teacher & 53.8 \\
         EMA (mean teacher) & 63.5
    \end{tabular}
    \label{tab:teacher_update}
\end{table}

We compare other approaches to updating the teacher during self-training vs. using exponential moving average in \cref{tab:teacher_update}. We see that EMA significantly outperforms using a fixed teacher (\ie vanilla self-training, where pseudo-labels are generated once before training) as well as using the student as its own teacher without EMA.

\subsection{Example of (Un)Fair Comparisons}
\label{appendix:unfair}

\begin{figure}
    \centering
    \includegraphics[width=0.8\linewidth]{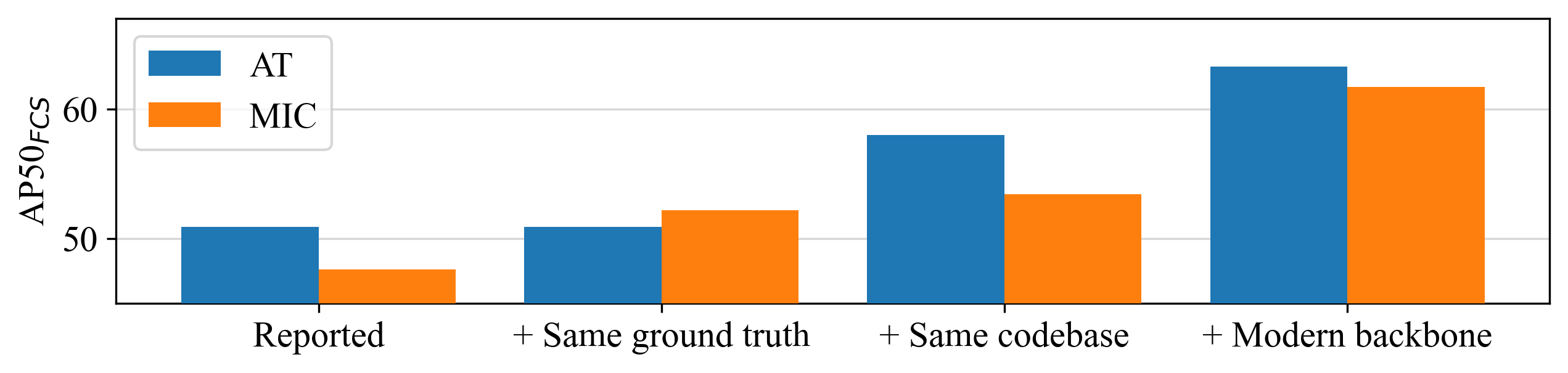}
    \vspace{-15pt}
    \caption{\textbf{Effects of fair and modernized comparison between MIC and AT.} Here we show an example of why fair and modern comparisons are necessary for making principled progress in DAOD. Moving left to right: \textbf{(1)} Published results report a difference of 3.3 AP50 on Cityscapes $\rightarrow$ Foggy Cityscapes between the two methods; \textbf{(2)} However the authors used \textit{different truth test labels}, and when this is corrected we see that the originally-published MIC model actually outperforms the originally-published AT model; \textbf{(3)} The authors also used different object detection libraries (Detectron2 for AT and maskrcnn-benchmark for MIC); when we re-implement them on top of ALDI (still using the VGG-16 backbones proposed in the original papers), we see that AT significantly outperforms MIC, but \textbf{(4)} These performance differences are less pronounced when using a modern backbone, indicating that for practical use there is less difference between these two methods than previously reported.}
    \label{fig:faircompare}
\end{figure}

In \cref{fig:faircompare} we show a case study of why fair comparisons are impactful for DAOD research. We compare two similarly-performing prior works, AT and MIC, and see that implementation inconsistencies have led to nontransparent comparisons between the two methods. Notably, the originally reported results even used different ground truth test labels. When re-implemented on top of the same modern framework using ALDI, we are able to fairly compare the two methods for the first time.

\subsection{Method-specific Source-Only Models}
\label{sec:method-specific}

\begin{figure}
    \centering
    \includegraphics[width=\linewidth]{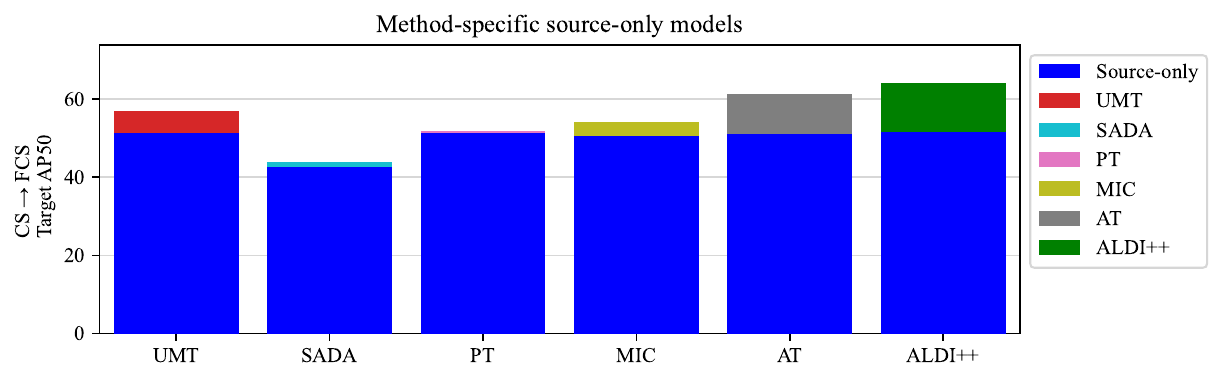}
    \caption{\textbf{Comparing methods to method-specific source-only models} that use the exact same training settings as the DAOD method in question. These results use ImageNet pre-training. Most of the bespoke source-only models perform very similarly, so in the main paper we only visualize one representative source-only model. The exception is SADA, whose corresponding source-only model performs worse due to the lack of EMA during training. See \cref{sec:method-specific}.}
    \vspace{-15pt}
    \label{fig:method-specific-baselines}
\end{figure}

Our protocol for training fair source-only models introduced in \cref{sec:baselinesoracles} is to utilize all techniques from the methods being studied that do not need target data to run. For simplicity, in the main paper we have only displayed the source-only model that utilizes the same components as ALDI++, though these settings differ slightly from the other methods studied. In \cref{fig:method-specific-baselines} we show an alternate view in which for each method we train a bespoke source-only model using the exact same training settings as the DAOD method. The main difference is the set of image augmentations used, except for SADA, which also does not use EMA. We see that there is only a small variation in the strength of the source-only models corresponding to each method, so our choice to only visualize one in the main paper for simplicity is reasonable. The exception is SADA, whose source-only model is significantly weakened by not using EMA.

\section{Implementation Details}
\label{sec:implementation}

\subsection{ALDI-YOLO}
\label{appendix:yolo}

We use an Detectron2 implementation of YOLOv5m as our starting point. All hyperparameter settings are identical to those of ALDI++. 

We implement soft distillation for YOLOv5 as follows. We compute an ``objectness'' (foreground/background) loss for each proposal, and compute classification and localization losses for pseudo-foreground labels only. Given pre-softmax student logits $l$ and teacher logits $\hat{l}$:

\begin{equation}
    L_{obj,soft} = BCE(l_{obj}, \hat{l}_{obj})
\end{equation}
\begin{equation}
    L_{cls,soft} = CE(l_{cls}, \alpha(\hat{l}_{cls})
\end{equation}
\begin{equation}
    L_{loc,soft} = CIOU(l_{loc}, \alpha(\hat{l}_{loc}))
\end{equation}
\begin{equation}
    L_{distill,soft} = L_{obj,soft} + L_{cls,soft} + L_{loc,soft}
\end{equation}

Where BCE is the binary cross-entropy loss, CE is the cross-entropy loss, CIOU is the Complete IoU loss~\cite{zheng2020distance}, and $\alpha$ is still a function of the post-softmax scores. See \cref{fig:comparison} for a visual depiction.

\subsection{ALDI-DETR}
\label{appendix:detr}

We use an Detectron2 implementation of Deformable DETR as our starting point. There is not an established technique for using soft knowledge distillation in end-to-end transformer-based queries like those of DETR; thus, we use hard distillation with a pseudo label threshold of 0.8. Similar to prior work~\cite{wang2021exploring,yu2022mttrans,jia2023pm}, we disable the EMA update for object query parameters.

\begin{table}[h]

\caption{\textbf{Settings to reproduce five prior works and our method \methodbest.} \textbf{Burn-in:} fixed duration (Fixed), our approach (Ours, \cref{sec:ours}). \textbf{Augs. }{\boldmath{$T_{src}, T_{tgt}$}}\textbf{:} Random flip (F), multi-scale (M), crop \& pad (CP), color jitter (J), gaussian blur (B), cutout~\citep{devries2017improved} (C), MIC~\citep{hoyer2023mic}. ${\frac{1}{2}}$: augs used on half the images in the batch. \textbf{\boldmath{$\frac{B_{tgt}}{B}$}:} Target-domain portion of minibatch of size $B$. \textbf{Postprocess:} Processing of teacher preds before distillation: sigmoid/softmax (Sharpen), sum class preds for pseudo-objectness (Sum), conf. thresholding (Thresh), NMS. \boldmath{$L_{distill}$}\textbf{:} Distillation losses: hard pseudo-labels (Hard), continuous targets (Soft). \boldmath{$L_{align}$}\textbf{:} Feature alignment losses: image-level adversarial (Img), instance-level adversarial (Inst), image-to-image translation (Img2Img). 
$^\dagger$: settings used in \methodname implementation (last column) but not in the original implementation (second-to-last column). $^{at}$: source-only and oracle results sourced from \cite{li2022cross}.
\label{tab:params}
}

\resizebox{\linewidth}{!}{

    \centering
    
    \begin{tabular}{c|c|c|c|r|c|c|c}
        \multirow{2}{*}{Method} & \cellcolor{Burnin}{$\theta_{stu}, \theta_{tch}$} & \cellcolor{T_src}  & \cellcolor{T_tgt}  & 
        \cellcolor{MtM} & \cellcolor{Postprocess}{Post-} & \cellcolor{L_distill} & \cellcolor{L_align}  \\
        
        & \cellcolor{Burnin}{Burn-in} & \cellcolor{T_src} \multirow{-2}{*}{$T_{src}$} & \cellcolor{T_tgt} \multirow{-2}{*}{{$T_{tgt}$}} & \cellcolor{MtM} \multirow{-2}{*}{$\frac{B_{tgt}}{B}$}& \cellcolor{Postprocess}{process} & \cellcolor{L_distill} \multirow{-2}{*}{{$L_{distill}$}} & \cellcolor{L_align} \multirow{-2}{*}{{$L_{align}$}}  \\
        
        \hline\hline

        \rule{0pt}{\normalbaselineskip}Source-only & -- & F, M$^\dagger$, C$^\dagger$, B$^\dagger$, E$^\dagger$ & -- & 0.0 & -- & -- & --  \\
         
        \hline

        \rule{0pt}{\normalbaselineskip}SADA~\citep{chen2021scale} & -- & F, M$^\dagger$ & F, M$^\dagger$ & 0.5 & -- & -- & Img, Inst  \\

         PT~\citep{chen2022learning} & Fixed & F, M$^\dagger$ & F, M$^\dagger$, J, C, B & 0.3 & Sharpen, Sum & Soft & --  \\

         UMT~\citep{deng2021unbiased} & -- & F$^\dagger$, M$^\dagger$ & F$^\dagger$, M$^\dagger$, CP, J, B & 0.5 & Thresh, NMS & Hard & Img2Img  \\

         MIC~\citep{hoyer2023mic} & -- & F, M$^\dagger$ & F, M$^\dagger$, J, B, MIC & 0.5 & Thresh, NMS & Hard & Img, Inst  \\

         AT~\citep{li2022cross} & Fixed & F, M$^\dagger$, J$^{\frac{1}{2}}$, C$^{\frac{1}{2}}$, B$^{\frac{1}{2}}$ & F, M$^\dagger$, J, C, B & 0.3 & Thresh, NMS & Hard & Img  \\
         
         \hline

         \rule{0pt}{\normalbaselineskip}\textbf{\methodbest} & Ours & F, M, J, C, B & F, M, J, B, MIC & 0.5 & Sharpen & Soft & --  \\
         
         \hline

         \rule{0pt}{\normalbaselineskip}Oracle & -- & -- & F, M$^\dagger$, J$^\dagger$, C$^\dagger$, B$^\dagger$ & 1.0 & -- & -- & -- \\
    \end{tabular}
}

\end{table}

\subsection{Re-implementations of Other Methods}

\begin{figure}
    \centering
    \includegraphics[width=\linewidth]{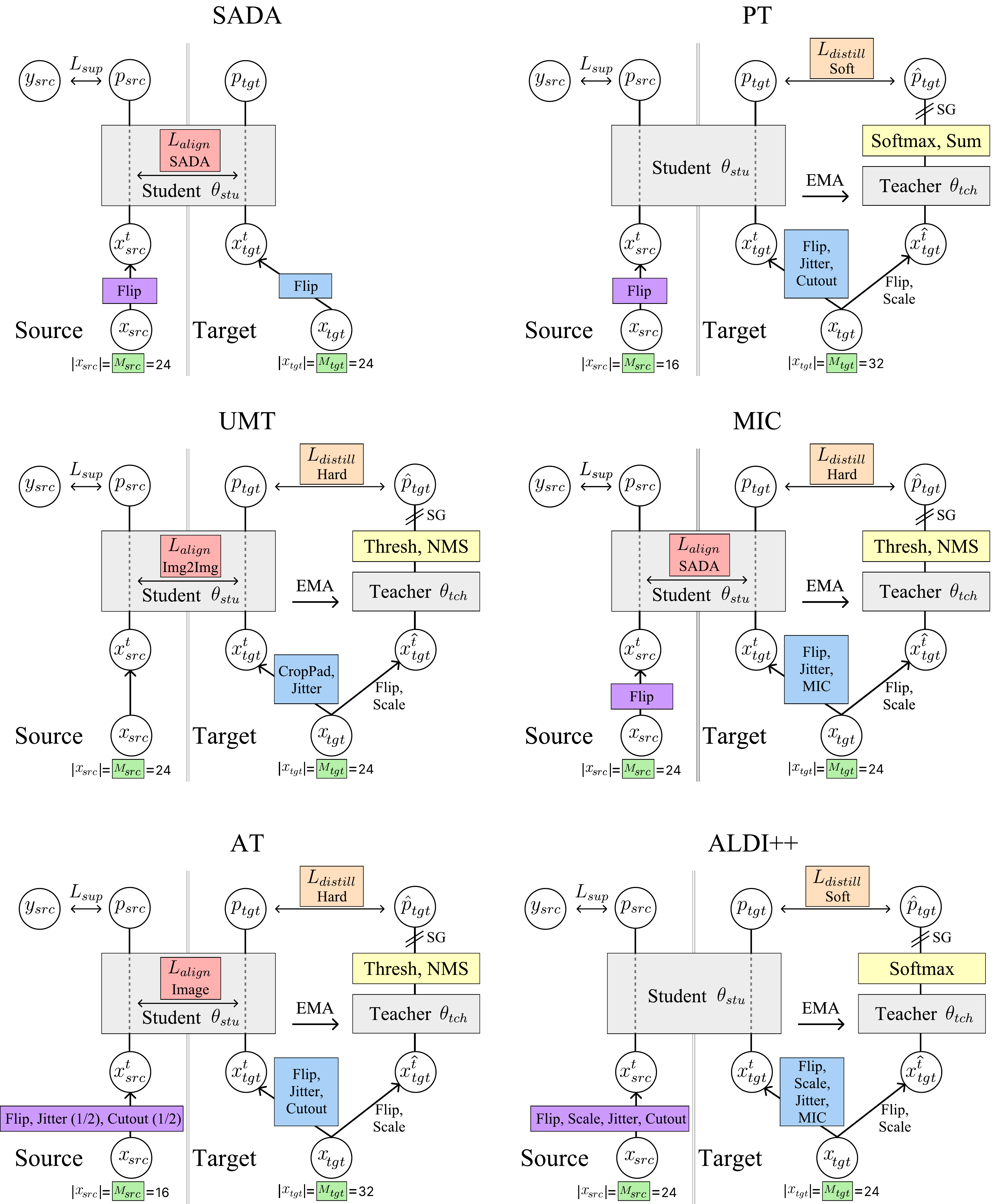}
    \caption{\textbf{Visual depiction of ALDI settings for reproducing prior work.}}
    \label{fig:allmethods}
\end{figure}

Here we include additional details regarding our re-implementations of prior work on top of the ALDI framework. All hyperparameters are reported in \cref{tab:params}. We visualize our implementations in \cref{fig:allmethods}.

\subsubsection{Adaptive Teacher \citep{li2022cross}}

Adaptive Teacher (AT) uses the default settings from the base configuration in Table 2 of the main paper, plus an image-level alignment network. For fair reproduction, we used the authors' alignment network implementation instead of our own for all AT experiments.

\subsubsection{MIC \citep{hoyer2023mic}}

We reimplemented the masked image consistency augmentation as a Detectron2 \texttt{Transform} in our framework for efficiency. We also implemented MIC's ``quality weight'' loss re-weighting procedure, though in our experiments we found that it makes performance slightly worse (AP50 on Foggy Cityscapes of 62.8 vs. 63.1 without).

\subsubsection{Probabilistic Teacher~\citep{chen2022learning}}

Probabilistic Teacher (PT) utilizes: (1) a custom Faster R-CNN architecture that makes localization predictions probabilistic, called ``Gaussian R-CNN'', (2) a focal loss objective, (3) learnable anchors. We ported implementations of these three components to our framework. Note that we first had to burn in a Gaussian R-CNN, so PT was not able to use the exact same starting weights as other methods. 

\subsubsection{SADA~\citep{chen2021scale}}

We port the official implementation of SADA to Detectron2. Note that SADA does not include burn-in or self-training, so the base implementation is the Detectron2 baseline config.

\subsubsection{Unbiased Mean Teacher~\citep{deng2021unbiased}}

Our implementation mirrors the UMT$_{SCA}$ configuration from ~\cite{deng2021unbiased}.

\subsection{Faster R-CNN Losses}
\label{appendix:fasterrcnn}

Here we describe the standard Faster R-CNN losses before describing how we modify them into ``soft'' distillation losses. Faster R-CNN consists of two stages: a region proposal network and the region-of-interest heads.

\subsubsection{Region proposal network (RPN):}
\label{sec:rpn}

\hspace{1pt}\\

\noindent
\textbf{Inputs.} The RPN takes as input:
\begin{enumerate}
    \item Features extracted by a backbone network (\eg a Resnet-50 with feature pyramid network in most of our experiments).
    \item A set of anchor boxes that represent the initial candidates for detection.
\end{enumerate}

\noindent
\textbf{Outputs.} For each anchor, the RPN predicts two things: 

\begin{enumerate}
    \item A binary classification called ``objectness'' representing whether the content inside the anchor box is foreground or background.
    \item Regression targets for the anchor, representing adjustments to the box to more closely enclose any foreground objects.
\end{enumerate}

\noindent
\textbf{Computing the loss.} In order to evaluate these predicted proposals, each proposal is matched to either foreground or background based on its intersection-over-union with the nearest ground truth box. Based on these matches, in the Detectron2 default settings a binary cross-entropy loss is computed for (1) and a smooth L1 loss is computed for (2).

A key challenge in Faster R-CNN is the severe imbalance between foreground and background anchors. To address this, a smaller number of proposals are sampled for computing the loss (256 in the default settings) with a specified foreground ratio (0.5 in the default setting). Objectness loss is computed for all proposals, while the box regression loss is computed only for foreground proposals (since it is undefined how the network should regress background proposals).

\subsubsection{Region of interest (ROI) heads:} 

\hspace{1pt}\\

\noindent
\textbf{Inputs.} The ROI heads take as input:
\begin{enumerate}
    \item Proposals from the RPN. In training, these are sampled at a desired foreground/background ratio, similar to the procedure used for computing the loss in the RPN. Note, however, that these will be different proposals than those used to compute RPN loss. In the Detectron2 defaults, 512 RPN proposals are sampled as inputs to the ROI heads at a foreground ratio of 0.25.
    \item Cropped backbone features, extracted using a procedure such as ROIAlign~\citep{he2017mask}. These are the features in the backbone feature map that are ``inside'' each proposal.
\end{enumerate}

\noindent
\textbf{Outputs.} The ROI heads then predict for each proposal:
\begin{enumerate}
    \item A multi-class classification.
    \item Regression targets for the final bounding box, representing adjustments to the box to more closely enclose any foreground objects.
\end{enumerate}

\noindent
\textbf{Computing the loss.} Predicted boxes are matched with ground truth boxes again based on intersection-over-union in order to compute the loss. By default we compute a cross-entropy loss for (1) and a smooth L1 loss for (2). (2) is again only computed for foreground predictions.

\subsection{Soft Distillation Losses for Faster R-CNN}

Distillation losses are computed between teacher predictions and student predictions. One option is select the teacher's most confident predictions based on a confidence threshold parameter to be ``pseudo-labels.'' These take the place of ground truth boxes in the standard Faster R-CNN losses for the student. We refer to this approach as using ``hard targets.''

In contrast, here we describe how we compute ``soft'' losses using intermediate outputs from the teacher to guide the student without thresholding. 

\noindent
\textbf{RPN.} The teacher and student RPNs start with the same anchors. We use the same sampling procedure described in \ref{sec:rpn} for choosing proposals for loss computation. Importantly, we ensure the \textit{same} proposals are sampled from the teacher and student so that they can be directly compared. We postprocess the teacher's objectness predictions with a sigmoid function to sharpen them. We then compute a binary cross-entropy loss between the teacher's post-sigmoid outputs and student's objectness predictions. We also compute a smooth L1 loss between the teacher's RPN regression predictions and the student's RPN regression predictions. Regression losses are only computed on proposals where the teacher's post-sigmoid objectness score is $\geq$ 0.8.

\noindent
\textbf{ROI heads.} The second stage of Faster R-CNN predicts a classification and regression for each RPN proposal; therefore, we need the input proposals to the student and teacher to be the same in order to directly compare their outputs. To achieve this, during soft distillation we initialize the student and teacher's ROI heads with the student's RPN proposals---intuitively, we want the teacher to tell the student ``what to do with'' its proposals from the first stage.

We postprocess the teacher's classification predictions with a softmax to sharpen them, then compute a cross-entropy loss between the teacher's post-softmax predictions and the student's classification predictions. We also compute a smooth L1 loss between the teacher's regression predictions and the student's regression predictions. We only compute regression losses where the teacher's top-scoring class prediction is not the background class.

\subsection{Adversarial Feature Alignment}
\label{sec:alignment_implementation}

We implement two networks to perform adversarial alignment at the image level and instance (bounding box) level. Our approach is inspired by Faster R-CNN in the Wild~\citep{chen2018domain} and SADA~\citep{chen2021scale}.

\noindent
\textbf{Image-level alignment.} We build an adversarial discriminator network that takes in backbone features at the image level. By default we use the ``p2'' layer of the feature pyramid network as described in~\cite{lin2017feature}. We use a simple convolutional head consisting of one hidden layer. Our defaults result in this \texttt{torch} module:

\footnotesize{
\begin{verbatim}
ConvDiscriminator(
    (model): Sequential(
      (0): Conv2d(256, 256, 
                    kernel_size=(3, 3), 
                    stride=(1, 1))
      (1): ReLU()
      (2): AdaptiveAvgPool2d(output_size=1)
      (3): Flatten(start_dim=1, end_dim=-1)
      (4): Linear(in_features=256, 
                    out_features=1, 
                    bias=True)
    )
  )
\end{verbatim}
}
\normalsize

\noindent
\textbf{Instance-level alignment.} We also implement an instance-level adversarial alignment network that takes as input the penultimate layer of the ROI heads classification head. By default, our instance level discriminator consists of one hidden fully-connected layer. Our defaults result in this \texttt{torch} module:

\footnotesize{
\begin{verbatim}
FCDiscriminator(
    (model): Sequential(
      (0): Flatten(start_dim=1, end_dim=-1)
      (1): Linear(in_features=1024, 
                    out_features=1024, 
                    bias=True)
      (2): ReLU()
      (3): Linear(in_features=1024, 
                    out_features=1, 
                    bias=True)
    )
)
\end{verbatim}
}
\normalsize

\section{Experiment Details}
\label{appendix:experiment_settings}

\subsection{Backbone Pretraining}
In our experiments, we evaluate two different backbones: a ResNet-50~\citep{he2016deep} with Feature Pyramid Network~\citep{lin2017feature}, and a ViT-B \citep{dosovitskiy2020image} with ViTDet~\citep{li2022exploring}. Both backbones are pre-trained on the ImageNet-1K classification and the COCO instance segmentation~\citep{lin2014microsoft} tasks. In addition, the ViT-B backbone is also pre-trained using the masked autoencoder objective proposed in~\cite{he2022masked}.

\subsection{Image-to-Image Translation}
\begin{figure*}[t]
    \centering
    \includegraphics[width=1\linewidth]{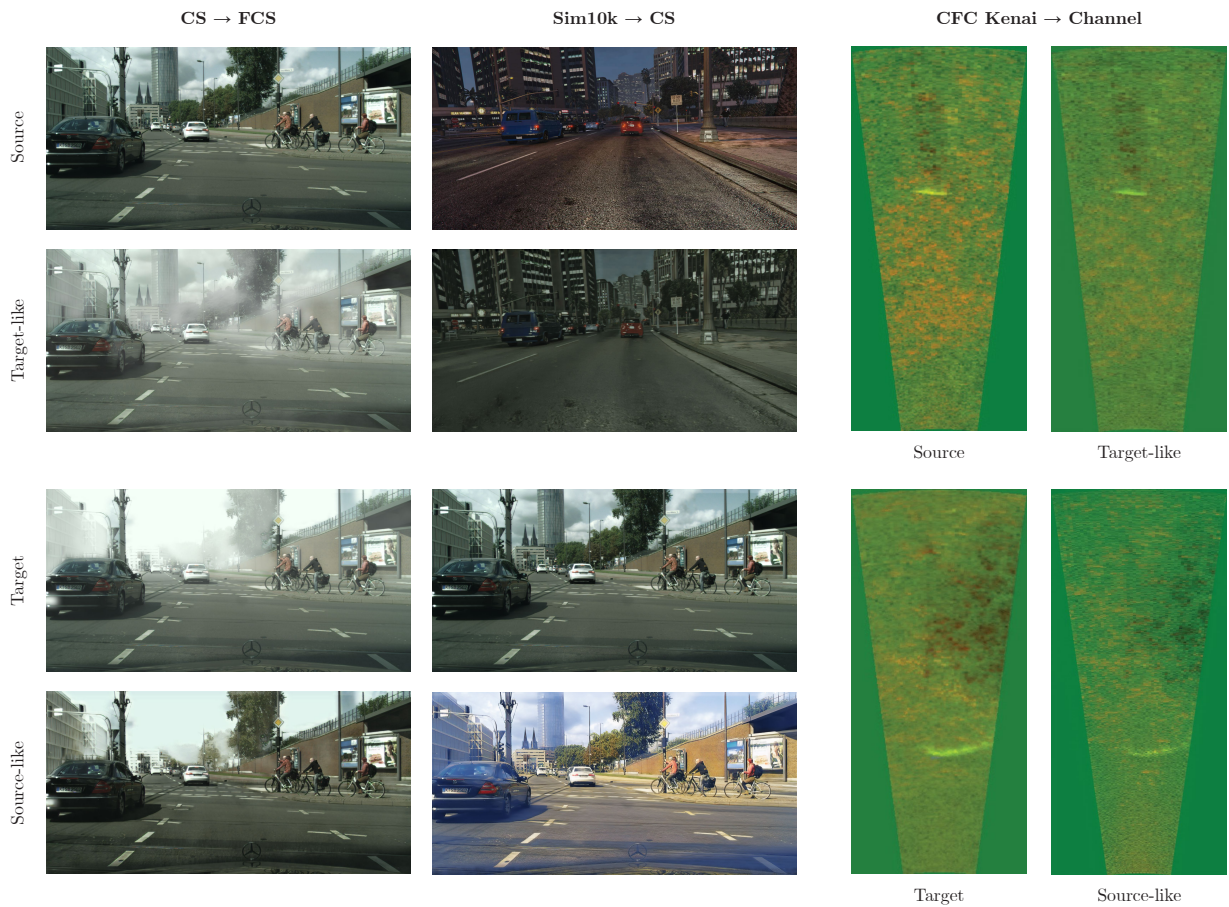}
    \caption{\textbf{Exemplary results of our CycleGAN models.} Source and target are the original images. Target-like and source-like are images translated by CycleGAN. Since FCS is derived from CS, CS $\rightarrow$ FCS is the only case in which we have paired images and can therefore show the translation from source into target-like and from target into source-like for the same example.}
    \label{fig:cyclegan_results}
\end{figure*}

In contrast to the adversarial alignment in feature space as in SADA \citep{chen2021scale}, UMT \citep{deng2021unbiased} aligns the domains in image (\ie pixel) space. This is achieved by training and using an unpaired image-to-image translation model to try to transform images from the source dataset into images that look like images from the target dataset (``target-like'') and vice-versa (``source-like''). We follow \cite{deng2021unbiased} by using the CycleGAN \citep{zhu2017unpaired} image-to-image translation model. We train the CycleGAN for 200 epochs (Cityscapes $\leftrightarrow$ Foggy Cityscapes, Sim10k $\leftrightarrow$ CS) or 20 epochs (Kenai $\leftrightarrow$ Channel) and respectively select the best model according to the average Fréchet inception distance (FID) \citep{heusel2017gans} between the source \& source-like and the target \& target-like images in the training dataset. For FID computation, we use the clean-fid implementation proposed in \cite{parmar2022aliased}. We compute FID on the training datasets as UMT only uses translated images thereof, which is why we are only interested in the best fit on the training data. We follow \cite{deng2021unbiased} by then generating source-like and target-like dataset using the selected model ahead of time, before the training of the main domain adaptation method. We note that 
tuning CycleGAN's hyperparameters or using other image-to-image translation methods could possibly improve UMT's performance however for the fair reproduction we use the defaults. We show some exemplary results of our CycleGAN models that are used to train UMT~\citep{deng2021unbiased} in Fig~\ref{fig:cyclegan_results}.

\subsection{Other Training Settings}
We fix the total effective batch size at 48 samples for fair comparison across all experiments. For training, we perform each experiments on 8 Nvidia V100 (32GB) GPUs distributed over four nodes. We use the MIT Supercloud~\citep{reuther2018interactive}.

\section{CFC-DAOD Dataset Details}
\label{appendix:cfc}

\noindent
Like other DAOD benchmarks, CFC-DAOD consists of data from two domains, source and target.

\subsection{Source data}

\noindent
\textbf{Train:} In CFC-DAOD, the source-domain training set consists of training data from the original CFC data release, i.e., video frames from the ``Kenai left bank'' location. We have used the 3-channel ``Baseline++'' format introduced in the original CFC paper~\citep{kay2022caltech}. For experiments in the ALDI paper, we subsampled empty frames to be around 10\% of the total data, resulting in 76,619 training images. For reproducibility, we release the exact subsampled set. When publishing results on CFC-DAOD, however, researchers are allowed to use the orignial CFC training set however they see fit and are not required to use our subsampled ``Baseline++'' data.

\noindent
\textbf{Validation:} The CFC-DAOD Kenai (source) validation set is the same as the original CFC validation set. We use the 3-channel ``Baseline++'' format from the original CFC paper. There are 30,454 validation images.

\subsection{Target data}

\noindent
\textbf{Train:} In CFC-DAOD, the target-domain ``training'' set consists of new data from the ``Kenai Channel'' location in CFC. These frames should be treated as unlabeled for DAOD methods, but labeled for Oracle methods. We also use the ``Baseline++'' format, and use the authors' original code for generating the image files from the original video files for consistency. There are 29,089 target train images.

\noindent
\textbf{Test:} The CFC-DAOD target-domain test set is the same as the ``Kenai Channel'' test set from CFC. We use the ``Baseline++'' format. There are 13,091 target test images. Researchers should publish final mAP@Iou=0.5 numbers on this data, and may use this data for model selection for fair comparison with prior methods.

\section{The ALDI Codebase}
\label{appendix:codebase}

We release ALDI as an open-source codebase built on a modern detector implementation. The codebase is optimized for speed, accuracy, and extensibility, training up to 5x faster than existing DAOD codebases while requiring up to 13x fewer lines of code. These qualities make our framework valuable for practitioners developing detection models in real applications, as well as for researchers pushing the state-of-the-art in DAOD. 

\begin{table}
    \centering
    \caption{\rule{0pt}{\normalbaselineskip}\textbf{Open-source codebases in domain adaptive object detection.} Existing methods use a variety of different detector implementations, including deprecated frameworks (maskrcnn-benchmark) and versions (Detectron2 $<$ v0.6). In contrast, \methodname is built on top of a modern framework, optimized for training speed, and is able to reproduce all five of these methods while requiring fewer lines of code (LOC) than any individual existing implementation. Our codebase can serve as a strong starting point for future research.}
    \begin{tabular}{c|c|c|c|r}
         \multirow{2}{*}{Codebase} & Faster R-CNN & \multirow{2}{*}{Backbone} & \multirow{2}{*}{Input size} & \multirow{2}{*}{LOC}  \\
         & Implementation  & & & \\
         \hline\hline
         \rule{0pt}{\normalbaselineskip}UMT~\citep{deng2021unbiased} & faster-rcnn.pytorch & VGG-16 & 600px & 19k \\
         SADA~\citep{chen2021scale} & maskrcnn-benchmark & Resnet50-FPN & 800px &  7k \\
         PT~\citep{chen2022learning} & Detectron2 v0.5 & VGG-16 & 600px &  3.4k \\
         MIC~\citep{hoyer2023mic} & maskrcnn-benchmark & Resnet50-FPN & 800px &  20k \\
         AT~\citep{li2022cross} & Detectron2 v0.3 & VGG-16 & 600px & 4k \\
         \hline
         \rule{0pt}{\normalbaselineskip}\methodname (Ours) & Detectron2 $\sim$v0.7 & Resnet50-FPN & 1024px & 1.5k \\
    \end{tabular}
    
    \label{tab:codebases}
\end{table}

\subsection{Detection Framework}

We designed the \methodname codebase to be lightweight and extensible. For this reason, we build on top of a recent version of Detectron2~\citep{wu2019detectron2}. The last tagged release of Detectron2 was \texttt{v0.6} in November 2021, however there have been a number upgrades since then leading to state-of-the-art performance. Thus, we use a fixed version that we call \texttt{v0.7ish} based off of an unofficial pull request for \texttt{v0.7}, commit \texttt{7755101} dated August 30 2023. We include this version of Detectron2 as a \texttt{pip}-installable submodule in the \methodname codebase for now, noting that once the official version is released it will no longer need to be a submodule (\ie it will be able to be directly installed through \texttt{pip} without cloning any code). 

Our codebase makes no modifications to the underlying Detectron2 code, making it a lightweight standalone framework. This is in contrast to existing DAOD codebases (see \cref{tab:codebases}) that often duplicate and modify the underlying framework as part of their implementation. By building \textit{on top of} Detectron2 rather than \textit{within} it, our codebase is up to 13x smaller than other DAOD codebases while providing more functionality. We note that in \cref{tab:codebases}, other codebases implement a single method while ours supports all methods studied.

\subsection{Speedups}

We found significant bottlenecks in training in other Detectron2-based codebases. Notably, we found that dataloaders and transform implementations were inefficient. These included, for instance:
\begin{itemize}
    \item Converting tensors back and forth between torch, numpy, and PIL during augmentation. We addressed this, reimplementing transforms as needed so that everything stays in torch.
    \item Using the random hue transform from torchvision. We found minimal changes in performance from disabling this component of the ColorJitter transform.
    \item Using separate dataloaders for weakly and strongly augmented imagery. We instead use a single dataloader per domain, with a hook to retrieve weakly augmented imagery before strong augmentations are performed.
\end{itemize}

\noindent
We reimplemented the dataloaders and augmentation strategies used by AT, MIC, and others to be more efficient, leading to a 5x speedup in training time per image compared to AT. 

\section{Tabular Results}
\label{appendix:tables}

\begin{table*}[h!]
  \caption{\textbf{Results with ResNet50-FPN backbones and ImageNet pre-training.}
  Previously-published results are shown in {\color{gray}gray}. Best results for each benchmark are in \textbf{bold} and second-best are \ul{underlined}.
  }

  \centering
  \label{table:city}
  \begin{tabular}{l|r|r|r}
  
  \toprule
  
  {Method} & CS $\rightarrow$ Foggy CS & Sim10k $\rightarrow$ CS & CFC Kenai $\rightarrow$ Channel\\

  \midrule
  
  Source-only & ({\color{gray}23.5}) 51.9 & ({\color{gray}35.5}) 61.4 & 67.3 \\

  \midrule
    UMT \citep{deng2021unbiased} & ({\color{gray}41.7}) 56.7 &  ({\color{gray}43.1}) 41.8 & 64.9\\
    
    SADA \citep{chen2021scale} & ({\color{gray}44.0}) 42.9 & 55.7 & 61.6\\
    
    PT \citep{chen2022learning} & ({\color{gray}47.1}) 51.5 &  ({\color{gray}55.1}) \ul{64.3} & 67.0\\
    
    MIC \citep{hoyer2023mic}  & ({\color{gray}47.6}) 54.0 &  \ul{64.3} & \ul{71.9}\\
    
    AT \citep{li2022cross}  & ({\color{gray}50.9}) \ul{61.0} &  56.7 & 69.2\\

    \midrule
    
    \textbf{ALDI++ (Ours)} & \textbf{63.9} & \textbf{69.1} & \textbf{72.6}\\
    
    \midrule
    	
    Oracle & ({\color{gray}42.7}) 62.4 & ({\color{gray}66.4}) 83.3 & 77.1 \\

  \bottomrule

  \end{tabular}
\end{table*}

\begin{table}[h!]
  \caption{\textbf{Results with ResNet50-FPN backbones and COCO pre-training.}
 Best results for each benchmark are in \textbf{bold} and second-best are \ul{underlined}.
  }

  \centering
  \label{table:city}
  \begin{tabular}{l|r|r|r}
  
  \toprule
  
  {Method} & CS $\rightarrow$ Foggy CS & Sim10k $\rightarrow$ CS & CFC Kenai $\rightarrow$ Channel\\

  \midrule
  
  Source-only & 59.1 & 76.8 & 66.7 \\

  \midrule
    UMT \citep{deng2021unbiased} & 61.4 & 58.7 & 61.2\\
    
    SADA \citep{chen2021scale} & 54.2 & 71.8 & 58.9\\
    
    PT \citep{chen2022learning} & 59.2 & 70.6 & 69.0\\
    
    MIC \citep{hoyer2023mic}  & 61.7 & \ul{73.1} & \ul{75.5}\\
    
    AT \citep{li2022cross} & \ul{63.3} & 72.0 & 69.1\\

    \midrule
    
    \textbf{ALDI++ (Ours)} & \textbf{66.8} & \textbf{77.8} & \textbf{76.1} \\
    
    \midrule
    	
    Oracle & 67.2 & 86.1 & 73.8 \\

  \bottomrule

  \end{tabular}
\end{table}

\begin{table}[h!]
  \caption{\textbf{Results with VitDet-B backbones and ImageNet pre-training.} Best results are in \textbf{bold} and second-best are \ul{underlined}.
  }

  \centering
  \label{table:city}
  \begin{tabular}{l|r}
  
  \toprule
  
  {Method} & CS $\rightarrow$ Foggy CS \\

  \midrule
  
  Source-only & 68.1 \\

  \midrule
    UMT \citep{deng2021unbiased} & 63.9 \\
    
    SADA \citep{chen2021scale} & 54.4 \\
    
    PT \citep{chen2022learning} & 65.8 \\
    
    MIC \citep{hoyer2023mic}  & \ul{70.6} \\
    
    AT \citep{li2022cross}  & 66.6 \\

    \midrule
    
    \textbf{ALDI++ (Ours)} & \textbf{71.0} \\
    
    \midrule
    	
    Oracle & 74.5 \\

  \bottomrule

  \end{tabular}
\end{table}

\section{Qualitative Results}
\label{sec:qualitative}

\begin{figure}[h]
    \centering
    \includegraphics[width=\linewidth]{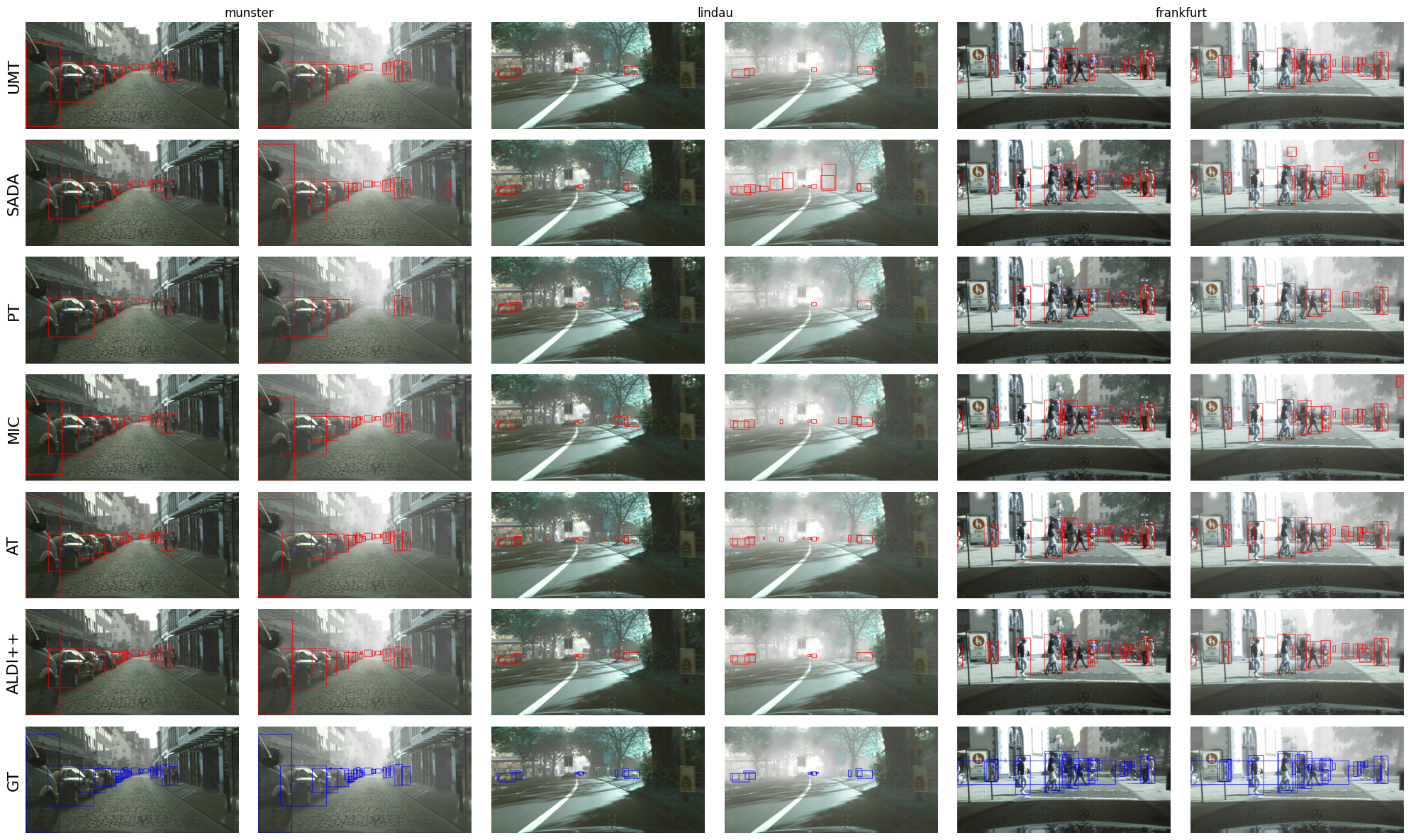}
    \caption{Qualitative results on Foggy Cityscapes. Best viewed maginified.}
    \label{fig:cityscapes-qual}
\end{figure}

\begin{figure}[h]
    \centering
    \includegraphics[width=\linewidth]{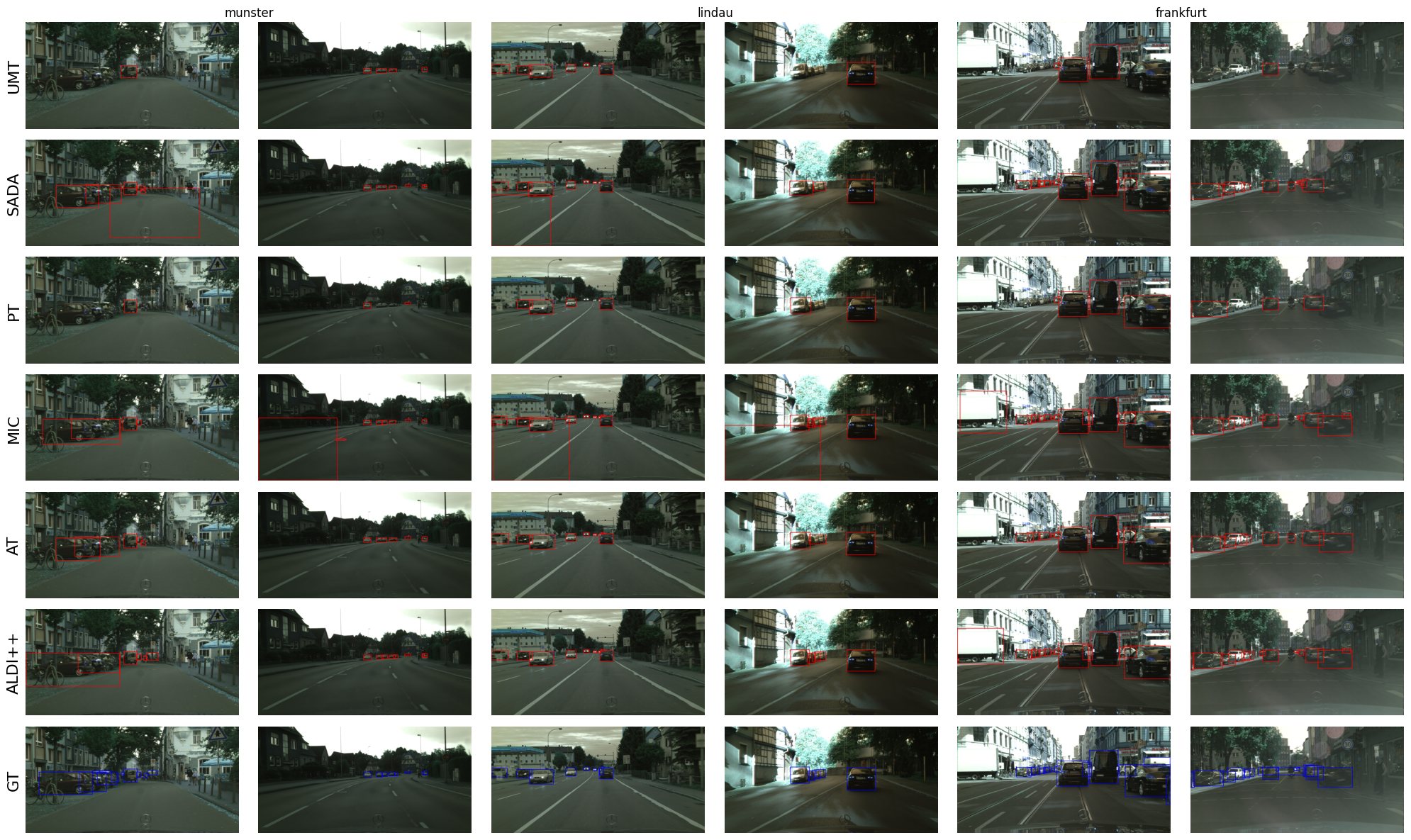}
    \caption{Qualitative results on Sim10k $\rightarrow$ Cityscapes. Best viewed maginified.}
    \label{fig:sim10k-qual}
\end{figure}

\begin{figure}[h]
    \centering
    \includegraphics[width=0.68\linewidth]{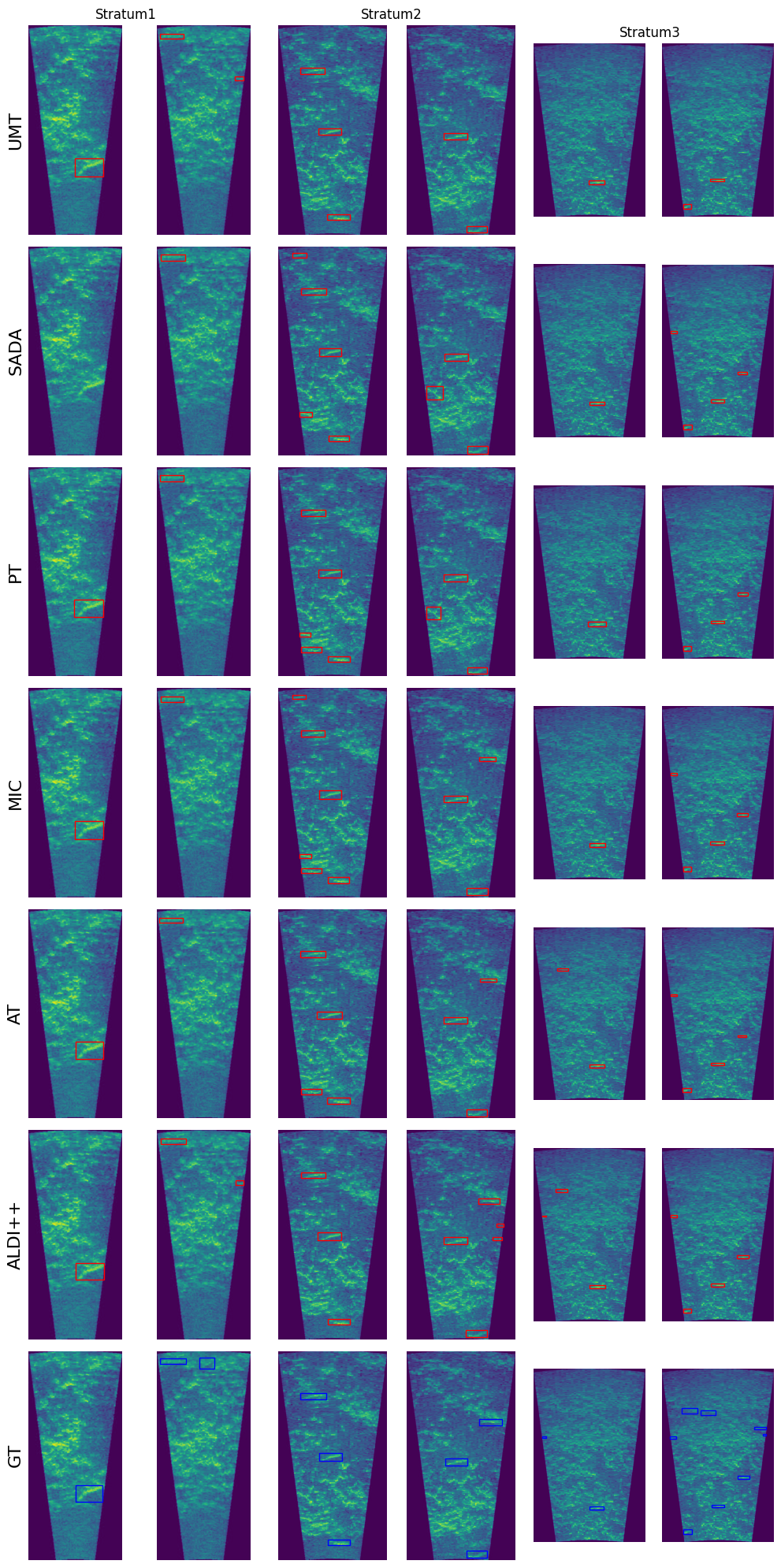}
    \caption{Qualitative results on the CFC-DAOD test set. Best viewed maginified.}
    \label{fig:cfc-qual}
\end{figure}

We visualize predictions from all models on Cityscapes $\rightarrow$ Foggy Cityscapes in \cref{fig:cityscapes-qual}, Sim10k $\rightarrow$ Cityscapes in \cref{fig:sim10k-qual}, and CFC-DAOD in \cref{fig:cfc-qual}. 

\textbf{Cityscapes $\rightarrow$ Foggy Cityscapes} We choose 1 random frame from each city in the validation set (Munster, Lindau, and Frankfurt) and show the 0.005 and 0.02 fog levels. We see that, compared to ALDI++, UMT and PT consistently suffer from more false positive detections, while other methods display both false positive and false negative detections. AT and MIC perform qualitatively similary to ALDI++ in many cases. AT suffers from more false positives than ALDI++ in all locations but fewer false negatives in Lindau. MIC performs better than ALDI++ in this randomly-selected Lindau frame. In the Frankfurt example, we see one downside of the Foggy Cityscapes benchmark: since detection annotations are generated programatically from segmentations, strange false positives can exist in the ground truth. Our new dataset CFC-DAOD addresses this problem by focusing directly on object detection. 

\textbf{Sim10k $\rightarrow$ Cityscapes} In \cref{fig:sim10k-qual} we compare results of all methods on the Sim10k $\rightarrow$ Cityscapes benchmark on two random images from each location in the Cityscapes validation set. In Munster, we see that all methods, including ALDI++, struggle with differentiating overlapping cars in the first image. In the second image, ALDI and MIC outperform other methods in differentiating cars in groups with less overlap than the first example. In Lindau, most prior work exhibits significantly more false negatives than ALDI++ and AT. AT, however, merges detections of multiple cars, leading to false negatives as well. In Frankfurt, most methods again exhibit significant false negatives, while ALDI++ and MIC show small false positive detections. Across all locations, MIC and SADA predict large false positives caused by foreground street lines.

\textbf{CFC-DAOD} In \cref{fig:cfc-qual} we compare results of all methods on the CFC-DAOD Kenai $\rightarrow$ Channel benchmark. We select two random images from each camera view in the test set. These camera views represent different range windows of the sonar camera, with Stratum1 the nearest-range (leading to the highest resolution and largest fish), followed by Stratum2, and finally Stratum3 is the longest-range (lowest resolution and smallest fish). We can see that each Stratum exhibits its own challenges in differentiating fish from the background, dealing with low signal-to-noise ratios, and differences in target size. In Stratum1, all methods except SADA detect the easily identifiable large fish in the first frame, but all methods suffer from the same false negative in the second frame. ALDI++ and UMT predict the same false positive near the edge of the field of view. In Stratum2, all methods suffer from false positives caused by background texture, and UMT, SADA, and PT suffer false negatives. In Stratum3, all methods fail to detect the furthest-range fish which are highly occluded by background texture, but exhibit fewer false positives than the other strata.

\end{document}

%% file: math_commands.tex
\usepackage{amsmath,amsfonts,bm}

\def\eqref#1{equation~\ref{#1}}

\def\1{\bm{1}}

\DeclareMathAlphabet{\mathsfit}{\encodingdefault}{\sfdefault}{m}{sl}
\SetMathAlphabet{\mathsfit}{bold}{\encodingdefault}{\sfdefault}{bx}{n}













%% file: main.bbl
\begin{thebibliography}{69}
\providecommand{\natexlab}[1]{#1}
\providecommand{\url}[1]{\texttt{#1}}
\expandafter\ifx\csname urlstyle\endcsname\relax
  \providecommand{\doi}[1]{doi: #1}\else
  \providecommand{\doi}{doi: \begingroup \urlstyle{rm}\Url}\fi

\bibitem[Arpit et~al.(2022)Arpit, Wang, Zhou, and Xiong]{arpit2022ensemble}
Devansh Arpit, Huan Wang, Yingbo Zhou, and Caiming Xiong.
\newblock Ensemble of averages: Improving model selection and boosting performance in domain generalization.
\newblock \emph{Advances in Neural Information Processing Systems}, 35:\penalty0 8265--8277, 2022.

\bibitem[Bondi et~al.(2018)Bondi, Fang, Hamilton, Kar, Dmello, Choi, Hannaford, Iyer, Joppa, Tambe, et~al.]{bondi2018spot}
Elizabeth Bondi, Fei Fang, Mark Hamilton, Debarun Kar, Donnabell Dmello, Jongmoo Choi, Robert Hannaford, Arvind Iyer, Lucas Joppa, Milind Tambe, et~al.
\newblock Spot poachers in action: Augmenting conservation drones with automatic detection in near real time.
\newblock In \emph{Proceedings of the AAAI Conference on Artificial Intelligence}, volume~32, 2018.

\bibitem[Cai et~al.(2019)Cai, Pan, Ngo, Tian, Duan, and Yao]{cai2019exploring}
Qi~Cai, Yingwei Pan, Chong-Wah Ngo, Xinmei Tian, Lingyu Duan, and Ting Yao.
\newblock Exploring object relation in mean teacher for cross-domain detection.
\newblock In \emph{Proceedings of the IEEE/CVF Conference on Computer Vision and Pattern Recognition}, pp.\  11457--11466, 2019.

\bibitem[Cao et~al.(2023)Cao, Joshi, Gui, and Wang]{cao2023contrastive}
Shengcao Cao, Dhiraj Joshi, Liang-Yan Gui, and Yu-Xiong Wang.
\newblock Contrastive mean teacher for domain adaptive object detectors.
\newblock In \emph{Proceedings of the IEEE/CVF Conference on Computer Vision and Pattern Recognition}, pp.\  23839--23848, 2023.

\bibitem[Carion et~al.(2020)Carion, Massa, Synnaeve, Usunier, Kirillov, and Zagoruyko]{carion2020end}
Nicolas Carion, Francisco Massa, Gabriel Synnaeve, Nicolas Usunier, Alexander Kirillov, and Sergey Zagoruyko.
\newblock End-to-end object detection with transformers.
\newblock In \emph{European conference on computer vision}, pp.\  213--229. Springer, 2020.

\bibitem[Caron et~al.(2021)Caron, Touvron, Misra, J{\'e}gou, Mairal, Bojanowski, and Joulin]{caron2021emerging}
Mathilde Caron, Hugo Touvron, Ishan Misra, Herv{\'e} J{\'e}gou, Julien Mairal, Piotr Bojanowski, and Armand Joulin.
\newblock Emerging properties in self-supervised vision transformers.
\newblock In \emph{Proceedings of the IEEE/CVF international conference on computer vision}, pp.\  9650--9660, 2021.

\bibitem[Chen et~al.(2017)Chen, Choi, Yu, Han, and Chandraker]{chen2017learning}
Guobin Chen, Wongun Choi, Xiang Yu, Tony Han, and Manmohan Chandraker.
\newblock Learning efficient object detection models with knowledge distillation.
\newblock \emph{Advances in neural information processing systems}, 30, 2017.

\bibitem[Chen et~al.(2022)Chen, Chen, Yang, Song, Wang, Zhang, Yan, Qi, Zhuang, Xie, et~al.]{chen2022learning}
Meilin Chen, Weijie Chen, Shicai Yang, Jie Song, Xinchao Wang, Lei Zhang, Yunfeng Yan, Donglian Qi, Yueting Zhuang, Di~Xie, et~al.
\newblock Learning domain adaptive object detection with probabilistic teacher.
\newblock \emph{arXiv preprint arXiv:2206.06293}, 2022.

\bibitem[Chen et~al.(2020)Chen, Kornblith, Norouzi, and Hinton]{chen2020simple}
Ting Chen, Simon Kornblith, Mohammad Norouzi, and Geoffrey Hinton.
\newblock A simple framework for contrastive learning of visual representations.
\newblock In \emph{International conference on machine learning}, pp.\  1597--1607. PMLR, 2020.

\bibitem[Chen et~al.(2018)Chen, Li, Sakaridis, Dai, and Van~Gool]{chen2018domain}
Yuhua Chen, Wen Li, Christos Sakaridis, Dengxin Dai, and Luc Van~Gool.
\newblock Domain adaptive faster r-cnn for object detection in the wild.
\newblock In \emph{Proceedings of the IEEE conference on computer vision and pattern recognition}, pp.\  3339--3348, 2018.

\bibitem[Chen et~al.(2021)Chen, Wang, Li, Sakaridis, Dai, and Van~Gool]{chen2021scale}
Yuhua Chen, Haoran Wang, Wen Li, Christos Sakaridis, Dengxin Dai, and Luc Van~Gool.
\newblock Scale-aware domain adaptive faster r-cnn.
\newblock \emph{International Journal of Computer Vision}, 129\penalty0 (7):\penalty0 2223--2243, 2021.

\bibitem[Cordts et~al.(2016)Cordts, Omran, Ramos, Rehfeld, Enzweiler, Benenson, Franke, Roth, and Schiele]{cordts2016cityscapes}
Marius Cordts, Mohamed Omran, Sebastian Ramos, Timo Rehfeld, Markus Enzweiler, Rodrigo Benenson, Uwe Franke, Stefan Roth, and Bernt Schiele.
\newblock The cityscapes dataset for semantic urban scene understanding.
\newblock In \emph{Proceedings of the IEEE conference on computer vision and pattern recognition}, pp.\  3213--3223, 2016.

\bibitem[Deng et~al.(2021)Deng, Li, Chen, and Duan]{deng2021unbiased}
Jinhong Deng, Wen Li, Yuhua Chen, and Lixin Duan.
\newblock Unbiased mean teacher for cross-domain object detection.
\newblock In \emph{Proceedings of the IEEE/CVF Conference on Computer Vision and Pattern Recognition}, pp.\  4091--4101, 2021.

\bibitem[DeVries \& Taylor(2017)DeVries and Taylor]{devries2017improved}
Terrance DeVries and Graham~W Taylor.
\newblock Improved regularization of convolutional neural networks with cutout.
\newblock \emph{arXiv preprint arXiv:1708.04552}, 2017.

\bibitem[Dosovitskiy et~al.(2020)Dosovitskiy, Beyer, Kolesnikov, Weissenborn, Zhai, Unterthiner, Dehghani, Minderer, Heigold, Gelly, et~al.]{dosovitskiy2020image}
Alexey Dosovitskiy, Lucas Beyer, Alexander Kolesnikov, Dirk Weissenborn, Xiaohua Zhai, Thomas Unterthiner, Mostafa Dehghani, Matthias Minderer, Georg Heigold, Sylvain Gelly, et~al.
\newblock An image is worth 16x16 words: Transformers for image recognition at scale.
\newblock \emph{arXiv preprint arXiv:2010.11929}, 2020.

\bibitem[Everingham et~al.(2010)Everingham, Van~Gool, Williams, Winn, and Zisserman]{everingham2010pascal}
Mark Everingham, Luc Van~Gool, Christopher~KI Williams, John Winn, and Andrew Zisserman.
\newblock The pascal visual object classes (voc) challenge.
\newblock \emph{International journal of computer vision}, 88\penalty0 (2):\penalty0 303--338, 2010.

\bibitem[Ganin \& Lempitsky(2015)Ganin and Lempitsky]{ganin2015unsupervised}
Yaroslav Ganin and Victor Lempitsky.
\newblock Unsupervised domain adaptation by backpropagation.
\newblock In \emph{International conference on machine learning}, pp.\  1180--1189. PMLR, 2015.

\bibitem[Ganin et~al.(2016)Ganin, Ustinova, Ajakan, Germain, Larochelle, Laviolette, Marchand, and Lempitsky]{ganin2016domain}
Yaroslav Ganin, Evgeniya Ustinova, Hana Ajakan, Pascal Germain, Hugo Larochelle, Fran{\c{c}}ois Laviolette, Mario Marchand, and Victor Lempitsky.
\newblock Domain-adversarial training of neural networks.
\newblock \emph{The journal of machine learning research}, 17\penalty0 (1):\penalty0 2096--2030, 2016.

\bibitem[Gao et~al.(2022)Gao, Sagawa, Koh, Hashimoto, and Liang]{gao2022out}
Irena Gao, Shiori Sagawa, Pang~Wei Koh, Tatsunori Hashimoto, and Percy Liang.
\newblock Out-of-distribution robustness via targeted augmentations.
\newblock In \emph{NeurIPS 2022 Workshop on Distribution Shifts: Connecting Methods and Applications}, 2022.

\bibitem[Guan \& Liu(2021)Guan and Liu]{guan2021domain}
Hao Guan and Mingxia Liu.
\newblock Domain adaptation for medical image analysis: a survey.
\newblock \emph{IEEE Transactions on Biomedical Engineering}, 69\penalty0 (3):\penalty0 1173--1185, 2021.

\bibitem[He et~al.(2016)He, Zhang, Ren, and Sun]{he2016deep}
Kaiming He, Xiangyu Zhang, Shaoqing Ren, and Jian Sun.
\newblock Deep residual learning for image recognition.
\newblock In \emph{Proceedings of the IEEE conference on computer vision and pattern recognition}, pp.\  770--778, 2016.

\bibitem[He et~al.(2017)He, Gkioxari, Doll{\'a}r, and Girshick]{he2017mask}
Kaiming He, Georgia Gkioxari, Piotr Doll{\'a}r, and Ross Girshick.
\newblock Mask r-cnn.
\newblock In \emph{Proceedings of the IEEE international conference on computer vision}, pp.\  2961--2969, 2017.

\bibitem[He et~al.(2020)He, Fan, Wu, Xie, and Girshick]{he2020momentum}
Kaiming He, Haoqi Fan, Yuxin Wu, Saining Xie, and Ross Girshick.
\newblock Momentum contrast for unsupervised visual representation learning.
\newblock In \emph{Proceedings of the IEEE/CVF conference on computer vision and pattern recognition}, pp.\  9729--9738, 2020.

\bibitem[He et~al.(2022)He, Chen, Xie, Li, Doll{\'a}r, and Girshick]{he2022masked}
Kaiming He, Xinlei Chen, Saining Xie, Yanghao Li, Piotr Doll{\'a}r, and Ross Girshick.
\newblock Masked autoencoders are scalable vision learners.
\newblock In \emph{Proceedings of the IEEE/CVF conference on computer vision and pattern recognition}, pp.\  16000--16009, 2022.

\bibitem[Heusel et~al.(2017)Heusel, Ramsauer, Unterthiner, Nessler, and Hochreiter]{heusel2017gans}
Martin Heusel, Hubert Ramsauer, Thomas Unterthiner, Bernhard Nessler, and Sepp Hochreiter.
\newblock Gans trained by a two time-scale update rule converge to a local nash equilibrium.
\newblock \emph{Advances in neural information processing systems}, 30, 2017.

\bibitem[Hinton et~al.(2015)Hinton, Vinyals, and Dean]{hinton2015distilling}
Geoffrey Hinton, Oriol Vinyals, and Jeff Dean.
\newblock Distilling the knowledge in a neural network.
\newblock \emph{arXiv preprint arXiv:1503.02531}, 2015.

\bibitem[Hoyer et~al.(2023)Hoyer, Dai, Wang, and Van~Gool]{hoyer2023mic}
Lukas Hoyer, Dengxin Dai, Haoran Wang, and Luc Van~Gool.
\newblock Mic: Masked image consistency for context-enhanced domain adaptation.
\newblock In \emph{Proceedings of the IEEE/CVF Conference on Computer Vision and Pattern Recognition}, pp.\  11721--11732, 2023.

\bibitem[Inoue et~al.(2018)Inoue, Furuta, Yamasaki, and Aizawa]{inoue2018cross}
Naoto Inoue, Ryosuke Furuta, Toshihiko Yamasaki, and Kiyoharu Aizawa.
\newblock Cross-domain weakly-supervised object detection through progressive domain adaptation.
\newblock In \emph{Proceedings of the IEEE conference on computer vision and pattern recognition}, pp.\  5001--5009, 2018.

\bibitem[Jia et~al.(2023)Jia, Liu, Yang, Wu, Xie, and Zhang]{jia2023pm}
Peidong Jia, Jiaming Liu, Senqiao Yang, Jiarui Wu, Xiaodong Xie, and Shanghang Zhang.
\newblock Pm-detr: Domain adaptive prompt memory for object detection with transformers.
\newblock \emph{arXiv preprint arXiv:2307.00313}, 2023.

\bibitem[Jocher et~al.(2023)Jocher, Chaurasia, and Qiu]{yolov8_ultralytics}
Glenn Jocher, Ayush Chaurasia, and Jing Qiu.
\newblock Ultralytics yolov8, 2023.
\newblock URL \url{https://github.com/ultralytics/ultralytics}.

\bibitem[Johnson-Roberson et~al.(2016)Johnson-Roberson, Barto, Mehta, Sridhar, Rosaen, and Vasudevan]{johnson2016driving}
Matthew Johnson-Roberson, Charles Barto, Rounak Mehta, Sharath~Nittur Sridhar, Karl Rosaen, and Ram Vasudevan.
\newblock Driving in the matrix: Can virtual worlds replace human-generated annotations for real world tasks?
\newblock \emph{arXiv preprint arXiv:1610.01983}, 2016.

\bibitem[Kalluri et~al.(2023)Kalluri, Xu, and Chandraker]{kalluri2023gnet}
Tarun Kalluri, Wangdong Xu, and Manmohan Chandraker.
\newblock Geonet: Benchmarking unsupervised adaptation across geographies.
\newblock \emph{CVPR}, 2023.

\bibitem[Kay et~al.(2022)Kay, Kulits, Stathatos, Deng, Young, Beery, Van~Horn, and Perona]{kay2022caltech}
Justin Kay, Peter Kulits, Suzanne Stathatos, Siqi Deng, Erik Young, Sara Beery, Grant Van~Horn, and Pietro Perona.
\newblock The caltech fish counting dataset: A benchmark for multiple-object tracking and counting.
\newblock In \emph{European Conference on Computer Vision}, pp.\  290--311. Springer, 2022.

\bibitem[Kay et~al.(2023)Kay, Stathatos, Deng, Young, Perona, Beery, and Van~Horn]{kay2023unsupervised}
Justin Kay, Suzanne Stathatos, Siqi Deng, Erik Young, Pietro Perona, Sara Beery, and Grant Van~Horn.
\newblock Unsupervised domain adaptation in the real world: A case study in sonar video.
\newblock In \emph{NeurIPS 2023 Computational Sustainability: Promises and Pitfalls from Theory to Deployment}, 2023.

\bibitem[Koh et~al.(2021)Koh, Sagawa, Marklund, Xie, Zhang, Balsubramani, Hu, Yasunaga, Phillips, Gao, et~al.]{koh2021wilds}
Pang~Wei Koh, Shiori Sagawa, Henrik Marklund, Sang~Michael Xie, Marvin Zhang, Akshay Balsubramani, Weihua Hu, Michihiro Yasunaga, Richard~Lanas Phillips, Irena Gao, et~al.
\newblock Wilds: A benchmark of in-the-wild distribution shifts.
\newblock In \emph{International Conference on Machine Learning}, pp.\  5637--5664. PMLR, 2021.

\bibitem[Li et~al.(2022{\natexlab{a}})Li, Mao, Girshick, and He]{li2022exploring}
Yanghao Li, Hanzi Mao, Ross Girshick, and Kaiming He.
\newblock Exploring plain vision transformer backbones for object detection.
\newblock In \emph{European Conference on Computer Vision}, pp.\  280--296. Springer, 2022{\natexlab{a}}.

\bibitem[Li et~al.(2022{\natexlab{b}})Li, Dai, Ma, Liu, Chen, Wu, He, Kitani, and Vajda]{li2022cross}
Yu-Jhe Li, Xiaoliang Dai, Chih-Yao Ma, Yen-Cheng Liu, Kan Chen, Bichen Wu, Zijian He, Kris Kitani, and Peter Vajda.
\newblock Cross-domain adaptive teacher for object detection.
\newblock In \emph{Proceedings of the IEEE/CVF Conference on Computer Vision and Pattern Recognition}, pp.\  7581--7590, 2022{\natexlab{b}}.

\bibitem[Lin et~al.(2014)Lin, Maire, Belongie, Hays, Perona, Ramanan, Doll{\'a}r, and Zitnick]{lin2014microsoft}
Tsung-Yi Lin, Michael Maire, Serge Belongie, James Hays, Pietro Perona, Deva Ramanan, Piotr Doll{\'a}r, and C~Lawrence Zitnick.
\newblock Microsoft coco: Common objects in context.
\newblock In \emph{European conference on computer vision}, pp.\  740--755. Springer, 2014.

\bibitem[Lin et~al.(2017)Lin, Doll{\'a}r, Girshick, He, Hariharan, and Belongie]{lin2017feature}
Tsung-Yi Lin, Piotr Doll{\'a}r, Ross Girshick, Kaiming He, Bharath Hariharan, and Serge Belongie.
\newblock Feature pyramid networks for object detection.
\newblock In \emph{Proceedings of the IEEE conference on computer vision and pattern recognition}, pp.\  2117--2125, 2017.

\bibitem[Liu et~al.(2021)Liu, Ma, He, Kuo, Chen, Zhang, Wu, Kira, and Vajda]{liu2021unbiased}
Yen-Cheng Liu, Chih-Yao Ma, Zijian He, Chia-Wen Kuo, Kan Chen, Peizhao Zhang, Bichen Wu, Zsolt Kira, and Peter Vajda.
\newblock Unbiased teacher for semi-supervised object detection.
\newblock \emph{arXiv preprint arXiv:2102.09480}, 2021.

\bibitem[Liu et~al.(2022)Liu, Mao, Wu, Feichtenhofer, Darrell, and Xie]{liu2022convnet}
Zhuang Liu, Hanzi Mao, Chao-Yuan Wu, Christoph Feichtenhofer, Trevor Darrell, and Saining Xie.
\newblock A convnet for the 2020s.
\newblock In \emph{Proceedings of the IEEE/CVF conference on computer vision and pattern recognition}, pp.\  11976--11986, 2022.

\bibitem[Morales-Brotons et~al.(2024)Morales-Brotons, Vogels, and Hendrikx]{morales2024exponential}
Daniel Morales-Brotons, Thijs Vogels, and Hadrien Hendrikx.
\newblock Exponential moving average of weights in deep learning: Dynamics and benefits.
\newblock \emph{Transactions on Machine Learning Research}, 2024.

\bibitem[Musgrave et~al.(2021)Musgrave, Belongie, and Lim]{musgrave2021unsupervised}
Kevin Musgrave, Serge Belongie, and Ser-Nam Lim.
\newblock Unsupervised domain adaptation: A reality check.
\newblock \emph{arXiv preprint arXiv:2111.15672}, 2021.

\bibitem[Musgrave et~al.(2022)Musgrave, Belongie, and Lim]{musgrave2022benchmarking}
Kevin Musgrave, Serge Belongie, and Ser-Nam Lim.
\newblock Benchmarking validation methods for unsupervised domain adaptation.
\newblock \emph{arXiv preprint arXiv:2208.07360}, 2022.

\bibitem[Nguyen et~al.(2020)Nguyen, Tseng, and Shuai]{nguyen2020domain}
Dang-Khoa Nguyen, Wei-Lun Tseng, and Hong-Han Shuai.
\newblock Domain-adaptive object detection via uncertainty-aware distribution alignment.
\newblock In \emph{Proceedings of the 28th ACM international conference on multimedia}, pp.\  2499--2507, 2020.

\bibitem[Oza et~al.(2023)Oza, Sindagi, Sharmini, and Patel]{oza2023unsupervised}
Poojan Oza, Vishwanath~A Sindagi, Vibashan~Vishnukumar Sharmini, and Vishal~M Patel.
\newblock Unsupervised domain adaptation of object detectors: A survey.
\newblock \emph{IEEE Transactions on Pattern Analysis and Machine Intelligence}, 2023.

\bibitem[Parmar et~al.(2022)Parmar, Zhang, and Zhu]{parmar2022aliased}
Gaurav Parmar, Richard Zhang, and Jun-Yan Zhu.
\newblock On aliased resizing and surprising subtleties in gan evaluation.
\newblock In \emph{Proceedings of the IEEE/CVF Conference on Computer Vision and Pattern Recognition}, pp.\  11410--11420, 2022.

\bibitem[Pham et~al.(2022)Pham, Cho, Joshi, and Hegde]{pham2022revisiting}
Minh Pham, Minsu Cho, Ameya Joshi, and Chinmay Hegde.
\newblock Revisiting self-distillation.
\newblock \emph{arXiv preprint arXiv:2206.08491}, 2022.

\bibitem[Redmon et~al.(2016)Redmon, Divvala, Girshick, and Farhadi]{redmon2016lookonceunifiedrealtime}
Joseph Redmon, Santosh Divvala, Ross Girshick, and Ali Farhadi.
\newblock You only look once: Unified, real-time object detection, 2016.
\newblock URL \url{https://arxiv.org/abs/1506.02640}.

\bibitem[Ren et~al.(2015)Ren, He, Girshick, and Sun]{ren2015faster}
Shaoqing Ren, Kaiming He, Ross Girshick, and Jian Sun.
\newblock Faster r-cnn: Towards real-time object detection with region proposal networks.
\newblock \emph{Advances in neural information processing systems}, 28, 2015.

\bibitem[Reuther et~al.(2018)Reuther, Kepner, Byun, Samsi, Arcand, Bestor, Bergeron, Gadepally, Houle, Hubbell, Jones, Klein, Milechin, Mullen, Prout, Rosa, Yee, and Michaleas]{reuther2018interactive}
Albert Reuther, Jeremy Kepner, Chansup Byun, Siddharth Samsi, William Arcand, David Bestor, Bill Bergeron, Vijay Gadepally, Michael Houle, Matthew Hubbell, Michael Jones, Anna Klein, Lauren Milechin, Julia Mullen, Andrew Prout, Antonio Rosa, Charles Yee, and Peter Michaleas.
\newblock Interactive supercomputing on 40,000 cores for machine learning and data analysis.
\newblock In \emph{2018 IEEE High Performance extreme Computing Conference (HPEC)}, pp.\  1--6. IEEE, 2018.

\bibitem[Rodriguez et~al.(2011)Rodriguez, Laptev, Sivic, and Audibert]{rodriguez2011density}
Mikel Rodriguez, Ivan Laptev, Josef Sivic, and Jean-Yves Audibert.
\newblock Density-aware person detection and tracking in crowds.
\newblock In \emph{2011 International Conference on Computer Vision}, pp.\  2423--2430. IEEE, 2011.

\bibitem[RoyChowdhury et~al.(2019)RoyChowdhury, Chakrabarty, Singh, Jin, Jiang, Cao, and Learned-Miller]{roychowdhury2019automatic}
Aruni RoyChowdhury, Prithvijit Chakrabarty, Ashish Singh, SouYoung Jin, Huaizu Jiang, Liangliang Cao, and Erik Learned-Miller.
\newblock Automatic adaptation of object detectors to new domains using self-training.
\newblock In \emph{Proceedings of the IEEE/CVF Conference on Computer Vision and Pattern Recognition}, pp.\  780--790, 2019.

\bibitem[Saito et~al.(2019)Saito, Ushiku, Harada, and Saenko]{saito2019strong}
Kuniaki Saito, Yoshitaka Ushiku, Tatsuya Harada, and Kate Saenko.
\newblock Strong-weak distribution alignment for adaptive object detection.
\newblock In \emph{Proceedings of the IEEE/CVF Conference on Computer Vision and Pattern Recognition}, pp.\  6956--6965, 2019.

\bibitem[Sakaridis et~al.(2018)Sakaridis, Dai, and Van~Gool]{sakaridis2018semantic}
Christos Sakaridis, Dengxin Dai, and Luc Van~Gool.
\newblock Semantic foggy scene understanding with synthetic data.
\newblock \emph{International Journal of Computer Vision}, 126:\penalty0 973--992, 2018.

\bibitem[Schneider et~al.(2018)Schneider, Taylor, and Kremer]{schneider2018deep}
Stefan Schneider, Graham~W Taylor, and Stefan Kremer.
\newblock Deep learning object detection methods for ecological camera trap data.
\newblock In \emph{2018 15th Conference on computer and robot vision (CRV)}, pp.\  321--328. IEEE, 2018.

\bibitem[Simonyan \& Zisserman(2014)Simonyan and Zisserman]{simonyan2014very}
Karen Simonyan and Andrew Zisserman.
\newblock Very deep convolutional networks for large-scale image recognition.
\newblock \emph{arXiv preprint arXiv:1409.1556}, 2014.

\bibitem[Tarvainen \& Valpola(2017)Tarvainen and Valpola]{tarvainen2017mean}
Antti Tarvainen and Harri Valpola.
\newblock Mean teachers are better role models: Weight-averaged consistency targets improve semi-supervised deep learning results.
\newblock \emph{Advances in neural information processing systems}, 30, 2017.

\bibitem[Vs et~al.(2021)Vs, Gupta, Oza, Sindagi, and Patel]{vs2021mega}
Vibashan Vs, Vikram Gupta, Poojan Oza, Vishwanath~A Sindagi, and Vishal~M Patel.
\newblock Mega-cda: Memory guided attention for category-aware unsupervised domain adaptive object detection.
\newblock In \emph{Proceedings of the IEEE/CVF Conference on Computer Vision and Pattern Recognition}, pp.\  4516--4526, 2021.

\bibitem[Wang et~al.(2021)Wang, Cao, Zhang, He, Zha, Wen, and Tao]{wang2021exploring}
Wen Wang, Yang Cao, Jing Zhang, Fengxiang He, Zheng-Jun Zha, Yonggang Wen, and Dacheng Tao.
\newblock Exploring sequence feature alignment for domain adaptive detection transformers.
\newblock In \emph{Proceedings of the 29th ACM International Conference on Multimedia}, pp.\  1730--1738, 2021.

\bibitem[Weinstein et~al.(2021{\natexlab{a}})Weinstein, Gardner, Saccomanno, Steinkraus, Ortega, Brush, Yenni, McKellar, Converse, Lippitt, et~al.]{weinstein2021general}
Ben~G Weinstein, Lindsey Gardner, Vienna Saccomanno, Ashley Steinkraus, Andrew Ortega, Kristen Brush, Glenda Yenni, Ann~E McKellar, Rowan Converse, Christopher Lippitt, et~al.
\newblock A general deep learning model for bird detection in high resolution airborne imagery.
\newblock \emph{bioRxiv}, 2021{\natexlab{a}}.

\bibitem[Weinstein et~al.(2021{\natexlab{b}})Weinstein, Graves, Marconi, Singh, Zare, Stewart, Bohlman, and White]{weinstein2021benchmark}
Ben~G Weinstein, Sarah~J Graves, Sergio Marconi, Aditya Singh, Alina Zare, Dylan Stewart, Stephanie~A Bohlman, and Ethan~P White.
\newblock A benchmark dataset for canopy crown detection and delineation in co-registered airborne rgb, lidar and hyperspectral imagery from the national ecological observation network.
\newblock \emph{PLoS computational biology}, 17\penalty0 (7):\penalty0 e1009180, 2021{\natexlab{b}}.

\bibitem[Wu et~al.(2019)Wu, Kirillov, Massa, Lo, and Girshick]{wu2019detectron2}
Yuxin Wu, Alexander Kirillov, Francisco Massa, Wan-Yen Lo, and Ross Girshick.
\newblock Detectron2.
\newblock \url{https://github.com/facebookresearch/detectron2}, 2019.

\bibitem[Xue et~al.(2023)Xue, Yang, Rajaraman, Zamzmi, and Antani]{xue2023cross}
Zhiyun Xue, Feng Yang, Sivaramakrishnan Rajaraman, Ghada Zamzmi, and Sameer Antani.
\newblock Cross dataset analysis of domain shift in cxr lung region detection.
\newblock \emph{Diagnostics}, 13\penalty0 (6):\penalty0 1068, 2023.

\bibitem[Yu et~al.(2022)Yu, Liu, Wei, Zhou, Nakata, Gudovskiy, Okuno, Li, Keutzer, and Zhang]{yu2022mttrans}
Jinze Yu, Jiaming Liu, Xiaobao Wei, Haoyi Zhou, Yohei Nakata, Denis Gudovskiy, Tomoyuki Okuno, Jianxin Li, Kurt Keutzer, and Shanghang Zhang.
\newblock Mttrans: Cross-domain object detection with mean teacher transformer.
\newblock In \emph{European Conference on Computer Vision}, pp.\  629--645. Springer, 2022.

\bibitem[Zhou et~al.(2023)Zhou, Jiang, and Lu]{zhou2023ssda}
Huayi Zhou, Fei Jiang, and Hongtao Lu.
\newblock Ssda-yolo: Semi-supervised domain adaptive yolo for cross-domain object detection.
\newblock \emph{Computer Vision and Image Understanding}, 229:\penalty0 103649, 2023.

\bibitem[Zhu et~al.(2017)Zhu, Park, Isola, and Efros]{zhu2017unpaired}
Jun-Yan Zhu, Taesung Park, Phillip Isola, and Alexei~A Efros.
\newblock Unpaired image-to-image translation using cycle-consistent adversarial networks.
\newblock In \emph{Proceedings of the IEEE international conference on computer vision}, pp.\  2223--2232, 2017.

\bibitem[Zhu et~al.(2019)Zhu, Pang, Yang, Shi, and Lin]{zhu2019adapting}
Xinge Zhu, Jiangmiao Pang, Ceyuan Yang, Jianping Shi, and Dahua Lin.
\newblock Adapting object detectors via selective cross-domain alignment.
\newblock In \emph{Proceedings of the IEEE/CVF Conference on Computer Vision and Pattern Recognition}, pp.\  687--696, 2019.

\bibitem[Zhu et~al.(2020)Zhu, Su, Lu, Li, Wang, and Dai]{zhu2020deformable}
Xizhou Zhu, Weijie Su, Lewei Lu, Bin Li, Xiaogang Wang, and Jifeng Dai.
\newblock Deformable detr: Deformable transformers for end-to-end object detection.
\newblock \emph{arXiv preprint arXiv:2010.04159}, 2020.

\end{thebibliography}
